\newif\ifcomments
\newif\ifconf
\newif\ifoldnotes
\newif\ifICLR
\newif\ifICML

\conffalse
\ICLRfalse
\ICMLfalse
\commentsfalse
\oldnotesfalse

\ifconf
\ifICLR
\documentclass[table]{article} % For LaTeX2e
\usepackage{iclr2024_conference,times}

\usepackage{amsmath,amsfonts,bm}

\newcommand{\binaryaddition}{\texttt{Binary Addition}}

\newcommand{\duplicatestring}{\texttt{Duplicate String}}
\newcommand{\evenpairs}{\texttt{Even Pairs}}
\newcommand{\modulararithmeticbrackets}{\texttt{Modular Arithmetic}}
\newcommand{\modulararithmeticsimple}{\texttt{Modular Arithmetic (Simple)}}
\newcommand{\oddsfirst}{\texttt{Odds First}}
\newcommand{\paritycheck}{\texttt{Parity Check}}
\newcommand{\reversestring}{\texttt{Reverse String}}
\newcommand{\stackmanipulation}{\texttt{Stack Manipulation}}
\newcommand{\solveequation}{\texttt{Solve Equation}}
\newcommand{\computesqrt}{\texttt{Compute Sqrt}}
\newcommand{\missingduplicate}{\texttt{Missing Duplicate}}
\newcommand{\cyclenavigation}{\texttt{Cycle Navigation}}
\newcommand{\binarymultiplication}{\texttt{Binary Multiplication}}
\newcommand{\bucketsort}{\texttt{Bucket Sort}}

\newcommand{\push}{\textsc{push}}

\newcommand{\pop}{\textsc{pop}}
\newcommand{\noop}{\textsc{no-op}}
\newcommand{\writeleft}{\textsc{write-left}}
\newcommand{\writeright}{\textsc{write-right}}
\newcommand{\writestay}{\textsc{write-stay}}
\newcommand{\jumpleft}{\textsc{jump-left}}
\newcommand{\jumpright}{\textsc{jump-right}}

\def\eqref#1{equation~\ref{#1}}
\def\1{\bm{1}}

\DeclareMathAlphabet{\mathsfit}{\encodingdefault}{\sfdefault}{m}{sl}
\SetMathAlphabet{\mathsfit}{bold}{\encodingdefault}{\sfdefault}{bx}{n}

\DeclareMathOperator*{\argmax}{arg\,max}

\newcommand{\cX}{\mathcal{X}}

\usepackage{xcolor}
\usepackage{hyperref}       % hyperlinks

\usepackage{graphicx}
 \hypersetup{
     colorlinks=true,
     linkcolor=teal,
     filecolor=teal,
     citecolor=teal,      
     urlcolor=teal,
}
\usepackage{url}
\usepackage{amsmath}
\usepackage{amsfonts}
\usepackage{amssymb}
\usepackage{mathtools}
\usepackage{amsthm}
\usepackage{bbold}
\usepackage{lipsum}

\else \ifICML

\documentclass{article}

\usepackage{microtype}
\usepackage{graphicx}
\usepackage{hyperref}

\usepackage{icml2024}

\usepackage{amsmath}
\usepackage{amssymb}
\usepackage{mathtools}
\usepackage{amsthm}

\usepackage[capitalize,noabbrev]{cleveref}
\fi %iclr if
\fi % ICML if 

\else  % if no conf, i.e. if we want a tech report
\documentclass[11pt, a4paper, logo, copyright]{googledeepmind}
\usepackage[authoryear, sort&compress, round]{natbib}

\fi

\usepackage{booktabs}       % professional-quality tables
\usepackage{multirow}
\usepackage{subcaption}
\usepackage{thmtools}
\usepackage{thm-restate}
\usepackage{wrapfig}
\usepackage{hyperref}

\ifcomments
\newcommand{\warning}[1]{{\color{red}#1}}
\newcommand{\todo}[1]{{\color{red}#1}}
\newcommand{\jg}[1]{{\color{brown}[JG: #1]}}
\newcommand{\mh}[1]{{\color{blue}[MH: #1]}}
\newcommand{\ec}[1]{{\color{red}[EC: #1]}}
\newcommand{\ar}[1]{{\color{magenta}[AR: #1]}}
\newcommand{\lo}[1]{{\color{blue}[LO: #1]}}
\newcommand{\tg}[1]{{\color{purple}[TG: #1]}}
\newcommand{\XXX}{{\color{red}XXX}}
\else
\newcommand{\warning}[1]{}
\newcommand{\todo}[1]{}
\newcommand{\jg}[1]{}
\newcommand{\mh}[1]{}
\newcommand{\ec}[1]{}
\newcommand{\ar}[1]{}
\newcommand{\lo}[1]{}
\newcommand{\tg}[1]{}
\newcommand{\XXX}{}
\fi

\newtheorem{theorem}{Theorem}

\newtheorem{definition}[theorem]{Definition}
\newtheorem{remark}[theorem]{Remark}

\newcommand{\bfcode}[1]{\texttt{#1}}

\title{Learning Universal Predictors}

\ifconf
\ifICLR

\author{Jordi Grau-Moya \thanks{corresponding author \texttt{\{jordigrau\}@google.com}} \\
\And
Tim Genewein \\
\And Marcus Hutter \\
\And Laurent Orseau \\
\And Gregoire Deletang \\
\And
Elliot Catt \\
\And
Joel Veness\\
}
\else \ifICML
\icmltitlerunning{Learning Universal Predictors}

\fi
\fi

\else % no conf

\correspondingauthor{jordigrau@google.com}

\keywords{
 Kolmogorov-complexity, universal prediction, in-context learning 
}
\author[*,1]{Jordi Grau-Moya}
\author[*,1]{Tim Genewein}
\author[*,1]{Marcus Hutter}
\author[*,1]{Laurent Orseau}
\author[]{Gr\'egoire D\'eletang}
\author[]{Elliot Catt}
\author[]{Anian Ruoss}
\author[]{Li Kevin Wenliang}
\author[]{Christopher Mattern}
\author[]{Matthew Aitchison}
\author[]{Joel Veness}

\affil[*]{Equal contributions.}
\affil[1]{Google DeepMind, London, United Kingdom}
\fi

\begin{document}

%ifconf if

\begin{abstract}
  Meta-learning has emerged as a powerful approach to train neural networks to learn new tasks quickly from limited data. Broad exposure to different tasks leads to versatile representations enabling general problem solving. But, what are the limits of meta-learning? In this work, we explore the potential of amortizing the most powerful universal predictor, namely Solomonoff Induction (SI), into neural networks via leveraging  meta-learning to its limits.
  We use Universal Turing Machines (UTMs) to generate training data used to expose networks to a broad range of patterns. We provide theoretical analysis of the UTM data generation processes and meta-training protocols. We conduct comprehensive experiments with neural architectures (e.g. LSTMs, Transformers) and algorithmic data generators of varying complexity and universality. Our results suggest that UTM data is a valuable resource for meta-learning, and that it can be used to train neural networks capable of learning universal prediction strategies. 
%   \iftrue\else \\ {\bf Keywords:} % \ifconf
%   universal prediction, neural networks, Solomonoff induction, CTW, Chomsky hierarchy, in-context learning, Turing machines, Transformers, LSTM, meta-learning
%   \fi 
\end{abstract}

\ifconf
\else
\maketitle
\fi

\ifICML \else
% \begin{figure}
%     \centering
%     \includegraphics[scale=0.22]{assets/solomonoff_summary_h.png}
%     \caption{Summary of our meta-learning methodology.}\label{fig:summary}
% \end{figure}
\begin{wrapfigure}{r}{0.47\textwidth}
    \includegraphics[width=0.47\textwidth]{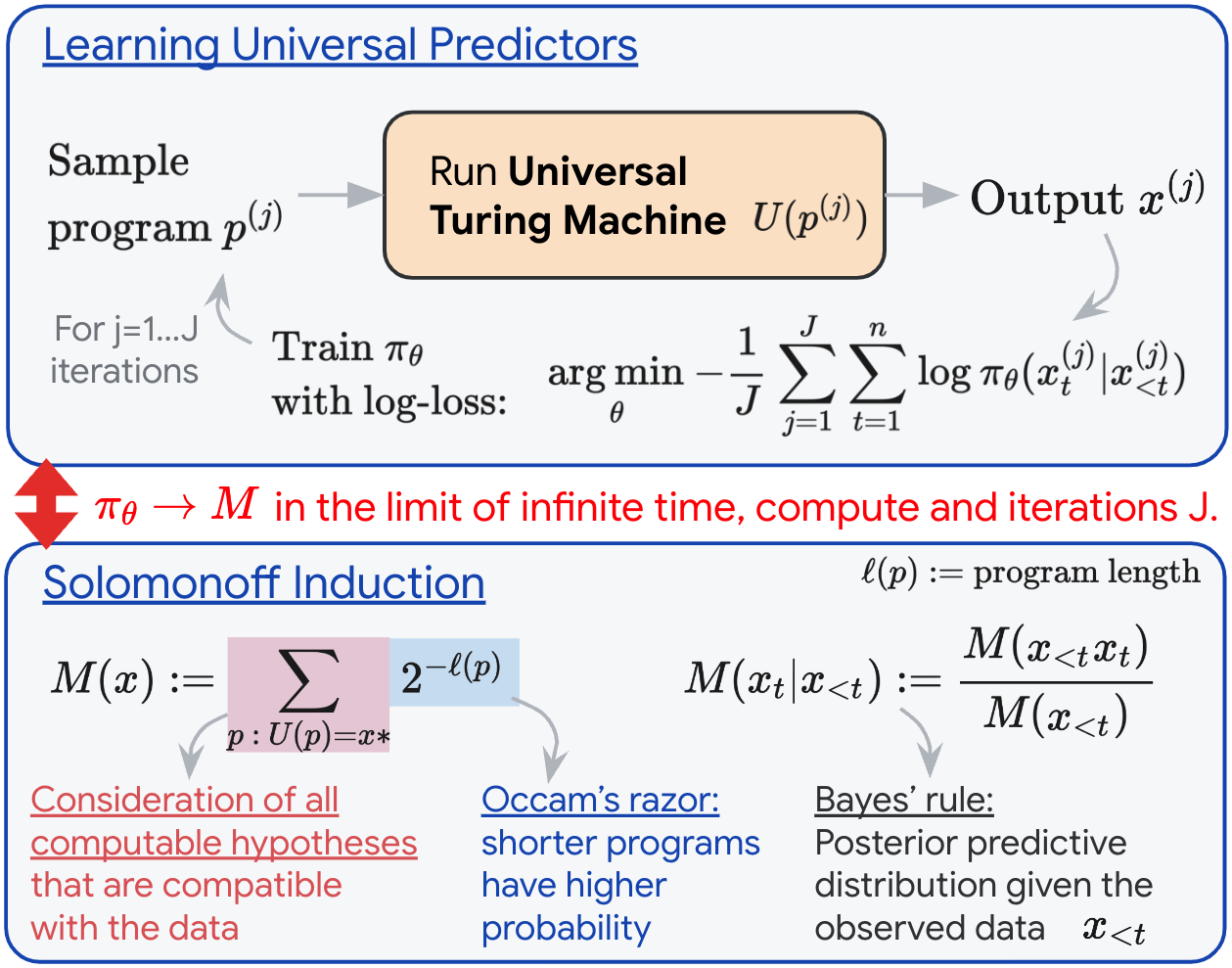}
    \caption{Summary of our meta-learning methodology.}\label{fig:summary}
    \vspace{3mm}
\end{wrapfigure}
\fi
Meta-learning has emerged as a powerful approach to enable AI systems to learn new tasks quickly from limited data~\citep{hospedales2021meta}. By training a model on a diverse set of tasks, meta-learning encourages the discovery of representations and learning strategies that generalize to new, unseen tasks. Intriguingly, recent research has shown that, when exposed to specific data regimes, meta-learning allows neural networks to perform Bayesian inference~\citep{ortega2019meta,mikulik2020meta,genewein2023memorybased}, which is critical for principled prediction under uncertainty. A key challenge in meta-learning is to design task distributions that are sufficiently broad, exposing the model to a rich variety of structures and patterns. Such broad exposure could lead to ``universal'' representations, enabling the system to tackle a wide range of problems and bringing us closer to the goal of artificial general intelligence (AGI).

\ifICML
\begin{figure}
\center
    \includegraphics[scale=0.17]{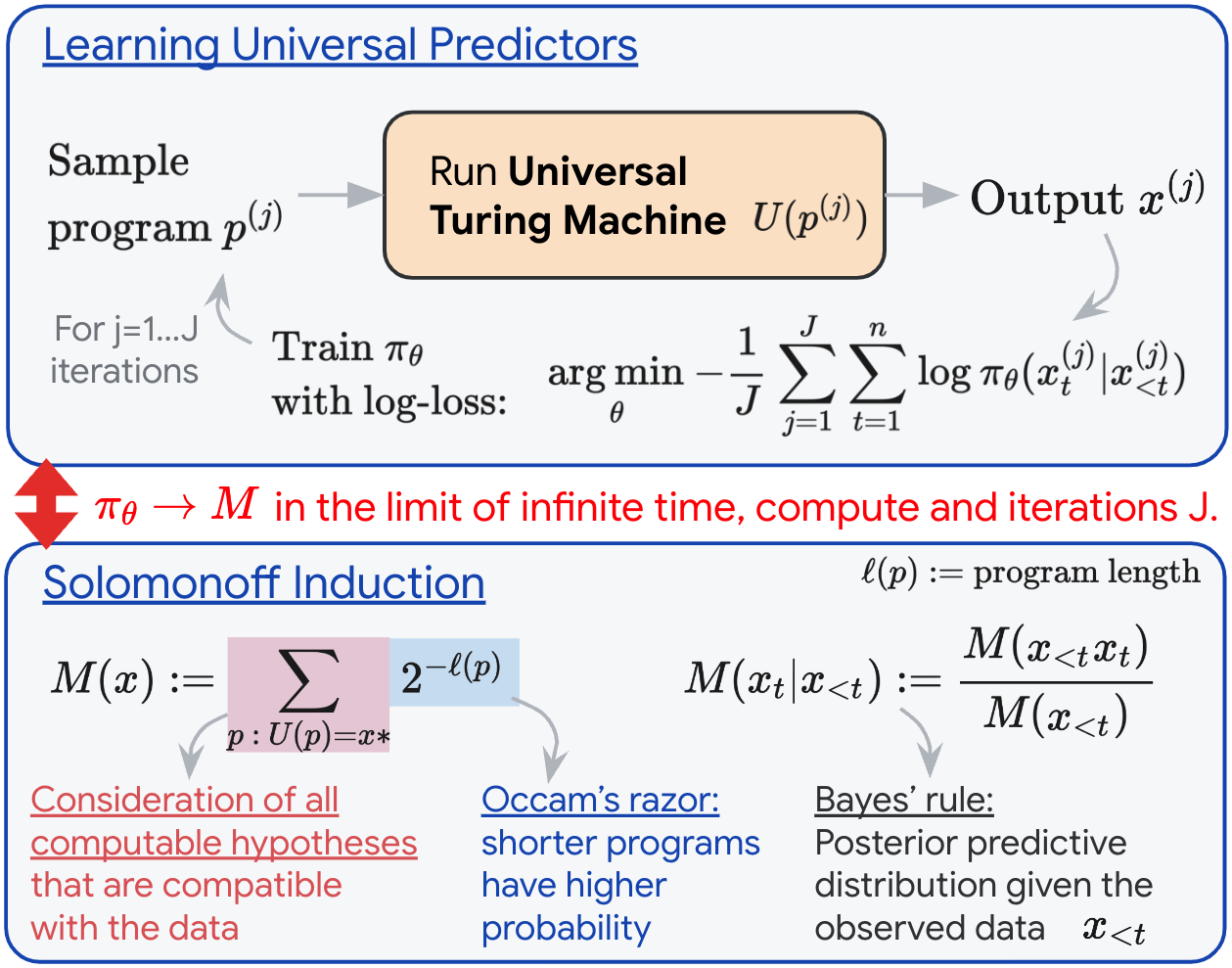}
    \caption{Summary of our meta-learning methodology.}
    \label{fig:summary}
\end{figure}
\fi

Solomonoff Induction
\footnote{
 SI arguably solved the century-old induction problem~\citep{rathmanner2011philosophical}, is 
 the basis of the Hutter prize~\citep{Hutter:06hprize}
and has been praised by the father of AI, Marvin Minsky: ``the most important discovery since Gödel''.}
(SI)  offers a compelling theoretical foundation for constructing such an ideal universal prediction system~\citep{solomonoff1964formal,solomonoff1964formal2}
\footnote{For an introduction see \citep{Hutter:07algprob,Hutter:17unilearn}  and see \citep{Hutter:07uspx} for technical details.} .
At its core, SI elegantly integrates three fundamental principles (see Figure~\ref{fig:summary}).
\emph{Consideration of all computable hypotheses:} Unlike traditional approaches, SI explores the entire space of computable hypotheses (i.e. generated by a computer program) as potential explanations for observed data.
\emph{Occam's Razor:} SI assigns higher prior probabilities to simpler hypotheses with shorter descriptions.
\emph{Bayesian Updating}: With new data, SI employs Bayes' rule to refine its belief about each hypothesis.
The theoretical strength of SI lies in its ability to rapidly converge on the true data-generating process, if computable~\citep{li1992inductive,hutter2004universal,sunehag2013principles,li2019introduction}. Yet, a significant barrier is its practical incomputability. The exhaustive exploration of algorithmic hypotheses demands immense computational resources.
To address this, approximations of SI were developed e.g. the Speed Prior~\citep{Schmidhuber:02speed,filan2016loss} and the Context Tree Weighting algorithm~\citep{willems1995context,willems1998context,veness2012ensemble}.

To understand the power of SI, imagine a program that generates an infinite stream of data $x$, e.g.,\ a fluid dynamics simulation or an AI movie generator. Let's say the length of the shortest possible version of this program  (i.e. its Kolmogorov complexity~\citep{li2019introduction})  is $N$ bits long, where all unnecessary elements have been removed and we have used compression to further reduce the size.  Now, if we feed the data stream $x$ to SI and let it predict each bit, something remarkable happens: After making fewer than $N$ prediction errors, SI will predict future data perfectly! This occurs because SI effectively learns the underlying rules of the data-generating program. With each incorrect prediction, it eliminates a range of possible explanations, allowing it to quickly find the correct program behind the data.

In this paper, we explore the potential of amortizing Solomonoff Induction into neural networks via meta-learning (see Figure~\ref{fig:summary}). A key challenge is finding neural architectures and training data distributions that guide networks towards learning SI in the limit. 
While neural networks are theoretically capable of universal computation~\citep{chen2017recurrent,stogin2020provably,mali2023computational}, practical training methods (e.g., stochastic gradient descent) can limit this ability~\citep{deletang2022neural}. Here we simply use off-the-shelf architectures like Transformers and LSTMs, while focusing on designing a suitable data training protocol. To address this, we generate data from  Universal Turing Machines (UTMs), which are fully general computers. Training on this ``universal data'' exposes the network to a broad space of computable patterns that guide the network towards learning universal inductive strategies.  
% We also test our networks on simpler (and not universal) algorithmic data  sources  (variable-order  Markov  and  tasks linked to the Chomsky hierarchy) for comparison and analysis.

\textbf{Our key contributions are:}
\emph{1) UTM data:} We use, for the first time, UTM data to meta-train neural networks.
\emph{2) Theoretical Analysis:} We provide a theoretical analysis of the UTM data generation process and training protocol that converges to SI in the limit.  
% We present a novel proof of the universality of non-uniform program distributions, crucial for efficient UTM data generation.
\emph{3) Extensive Experiments:} We conduct comprehensive experiments with a variety of neural architectures (e.g. LSTMs, Transformers) and algorithmic data generators of varying complexity and universality. 
We open-sourced the generators at \url{https://github.com/google-deepmind/neural_networks_solomonoff_induction}.
% \footnote{Open-sourced generators~\url{https://github.com/google-deepmind/neural_networks_solomonoff_induction}} . 

Our results show that increasing model size leads to improved performance, demonstrating that model scaling helps learning increasingly universal prediction strategies.
We find that: Large Transformers trained on UTM data successfully transfer their learning to other tasks suggesting they acquired reusable universal patterns;  On variable-order Markov sources, large LSTMs and Transformers achieve optimal performance, highlighting their ability to model Bayesian mixtures over programs necessary for SI.

%%%%%%%%%%%%%%%%%%%%%%%%%%%%%%%%%%%%%%%%%%%%%%%%%%%%%%%%%%%%%%%
\section{Background}\label{sec:background}
%%%%%%%%%%%%%%%%%%%%%%%%%%%%%%%%%%%%%%%%%%%%%%%%%%%%%%%%%%%%%%%

\textbf{Notation.}
An alphabet $\cX$ is a finite, non-empty set of symbols. 
A string $x_1x_2 \ldots x_n \in \cX^n$ of length $n$ is denoted by $x_{1:n}$.
The prefix $x_{1:j}$ of $x_{1:n}$, $j\leq n$, is denoted by $x_{\leq j}$ or $x_{< j+1}$.
The empty string is denoted by $\epsilon$.
Our notation generalizes to out-of-bounds indices i.e.\ given a string $x_{1:n}$ and an integer $m > n$, we define $x_{1:m} := x_{1:n}$ and $x_{n:m}:=\epsilon$.
The concatenation of two strings $s$ and $r$ is denoted by $sr$. The expression $[\![A ]\!]$ is $1$ if $A$ is true and $0$ otherwise.

\textbf{Semimeasures.}
A semimeasure is a probability measure $P$ over infinite and finite sequences $\cX^\infty\cup\cX^*$ 
for some finite alphabet $\cX$ assumed to be $\{0,1\}$
(most statements hold for arbitrary finite $\cX$).
Let $\mu(x)$ be the probability that an (in)finite sequence \emph{starts} with $x$.  While proper distributions satisfy $\sum_{a\in\cX}\mu(xa) = \mu(x)$, semimeasures exhibit \emph{probability gaps} and satisfy  $\sum_{a\in\cX}\mu(xa)\leq\mu(x)$.

\textbf{Turing Machines.} A Turing Machine (TM) takes a string of symbols $z$ as an input, and outputs a string of symbols $x$ (after reading $z$ and halting), i.e.\ $T(z) = x$. For convenience we define the output string at computation step $s$ as $T^s(z)=x$ which may be the empty string $\epsilon$. We adopt similar notation for Universal Turing Machines $U$. Monotone TMs (see Definition~\ref{def:monotonicity} below) are special TMs that can incrementally build the output string while incrementally reading the input program, which is a convenient practical property we exploit in  our experiments.
\begin{definition}[Monotonicity]\label{def:monotonicity}
 A universal machine $U$ is monotone if for all $p,q,x,y$ with $U(p)=y$ and $U(q)=x$ we have that $\ell(x)\ge \ell(y)$ and $p\sqsubseteq q$ imply $y\sqsubseteq x$, where $p\sqsubseteq q$ means that $p$ is a prefix string of $q$.  See Appendix~\ref{sec:generalized_solomonoff_proof} for a more thorough description.
\end{definition}

\textbf{Solomonoff Induction (SI).} 
The optimal prediction over the next symbol $x_{n+1}$ given an observed sequence $x_{1:n}$ is $\mu(x_{n+1} | x_{1:n}) = \mu(x_{1:n+1})/\mu(x_{1:n})$, assuming that $\mu$ is the true (but unknown) computable probability distribution over sequences. 
In contrast, SI predicts the next symbol $x_{n+1}$ using a single universal semimeasure $M$ widely known as the Solomonoff Universal Prior (see definition below).   
\begin{definition}[(Monotone) Solomonoff Prior] \label{def:monotone_solomonoff_prior}
Let $U$ be a universal monotone machine, then the Solomonoff prior is defined as
% \begin{equation}
$  M(x) ~:=~ \sum_{p:U(p)=x*} 2^{-\ell(p)}$
% \end{equation}
  with the sum is over all $p\in\{0,1\}^*$,
  where the output $x*$ is any string that starts with $x$ \emph{and} the whole program $p$ has been read by $U$.
%  \[ M(x) := M_{U}(x) = \sum_{p:~ x\in U_M(p)}2^{-\ell(p)} \]
\end{definition}
We can use $M$ to construct the posterior predictive distribution $M(x_{n+1} | x_{1:n}) = \frac{ M( x_{1:n}x_{n+1} ) }{ M( x_{1:n} ) }$ (see Figure~\ref{fig:summary}). This is equivalent to performing Bayesian inference on program space  $M(x_{n+1} | x_{1:n}) = \sum_{p} P(p | x_{1:n}) [\![ U(p) = x_{1:n}x_{n+1}*]\!]$ (for prefix-free programs, and any continuation $*$  of the sequence), where $P(p | x_{1:n})$ is the Bayesian posterior over programs given the data using the prior $P(p) = 2^{-\ell(p)}$ and the zero-one likelihood $P(x | p) = [\![ U(p) = x* ]\!]$.

% where $p$ is a program of length $\ell(p)$ and $U$ is a UTM that outputs  outputs $x$ when running program $p$.
\ifICML Solomonoff~\yrcite{solomonoff1964formal} \else \cite{solomonoff1964formal} \fi showed that $M$ converges fast (to the true $\mu$) if the data is generated by \emph{any} 
computable probability distribution $\mu$:
$\sum_{t=1}^\infty \sum_{x_{<t}} \mu(x_{<t}) \sum_{x\in \cX} ( M (x|x_{<t}) - \mu(x|x_{<t}) )^2 \le K(\mu)\ln 2 < \infty$,
% \begin{equation}
%     \sum_{t=1}^\infty \sum_{x_{<t}} \mu(x_{<t}) \sum_{x\in \cX} \left( \sqrt{M (x|x_{<t})} - \sqrt{\mu(x|x_{<t})} \right)^2 \le K(\mu)\ln 2 < \infty
% \end{equation}
where $K(\mu) := \min_p \{\ell (p) : U(p) = \mu\}$ is the Kolmogorov complexity~\citep{li2019introduction} of the generator $\mu$ (represented as a bitstring). This can be seen when noticing that on the left-hand-side of the inequality we have an infinite sum and on the right we have a constant.
The Solomonoff prior is essentially the best universal predictor given a choice of reference UTM.

There exists a normalized version of the Solomonoff prior (among others~\citep{wood2013non}) that is not a semimeasure but a proper measure i.e., properly normalized (see Definition~\ref{def:normalized_solomonoff_prior} below). It has nicer properties when $x$ contains incomputable sub-sequences~\citep{lattimore2011universal} and maintains the convergence properties of the standard Solomonoff prior. This version of SI is of interest to us because it suited to be learned by neural models (that are also properly normalized) and exhibits more efficient sampling than semimeasures (due to no probability gap).
\begin{definition}[Normalized Solomonoff Prior]\label{def:normalized_solomonoff_prior}
 For $a\in\cX$, Solomonoff normalization is defined as
 $  M^{norm}(\epsilon):=1,~~~ M^{norm}(a|x) 
  ~:=~ \frac{M(xa)}{\sum_{a\in\cX}M(xa)}
  ~=~ \frac{M^{norm}(xa)}{M^{norm}(x)} $.
\end{definition} 

\textbf{Algorithmic Data Generating Sources and the Chomsky Hierarchy.}
An algorithmic data generating source $\mu$ is
simply a computable data source by, for example, a TM $T$ fed with random inputs.
There is a natural hierarchy over machines based on their memory structure known as the Chomsky hierarchy (CH)~\citep{chomsky1956three}, which classifies sequence prediction problems---and associated automata models that solve them---by increasing complexity.
There are four levels in the CH, namely, regular, context-free, context-sensitive, and recursively enumerable. Solving problems on each level requires different memory structures such as finite states, stack, finite tape and infinite tape, respectively.
Note that any reasonable approximation to SI would need to sit at the top of the hierarchy.

\textbf{Meta-Learning.}
A parametric model $\pi_\theta$ can be meta-trained by repeating the following steps (see Figure~\ref{fig:summary}): 1) sample a task $\tau$ (programs in our case) from the task distribution $p(\tau)$, 2) sample an output sequence $x_{1:n}$ from $\tau$, 3) train the model $\pi_\theta$ with the log-loss $-\sum_{t=1}^n \log \pi_\theta (x_t | x_{<t})$. \ifICML Ortega et al.~\yrcite{ortega2019meta} \else \cite{ortega2019meta} \fi showed that the fully trained  $\pi_\theta$ behaves as a Bayes-optimal predictor, i.e.\  $\pi_\theta(x_{t}| x_{<t}) \approx \sum_\tau p(\tau |x_{<t}) p(x_t|x_{<t},\tau)$
where $p(x_t|x_{<t},\tau)$ is the predictive distribution,
and $p(\tau |x_{<t})$ the posterior~\citep{ortega2019meta}. More formally, if $\mu$ is a proper measure and $D=(x^1,...,x^J)$ are sequences cut to length $n$ sampled from $\mu$ 
with empirical distribution $\hat\mu(x)=\frac1J\sum_{y\in D}[\![y=x]\!]$,
then the log-loss
$ \text{Loss}(\theta) := -\frac1J\sum_{x\in D}\sum_{t=1}^{\ell(x)}\log \pi_\theta(x_t|x_{<t})
  = -\frac1J\sum_{x\in D}\log \pi_\theta(x) = -\sum_{x\in\cX^n} \hat\mu(x)\log p_\theta(x)$
is minimized for $\pi_\theta(x)=\hat\mu(x)$ provided $\pi_\theta$ can represent $\hat\mu$.

%%%%%%%%%%%%%%%%%%%%%%%%%%%%%%%%%%%%%%%%%%%%%%%%%%%%%%%%%%%%%%%
\section{Meta-Learning as an Approximation to Solomonoff Induction}\label{sec:MLASI}
%%%%%%%%%%%%%%%%%%%%%%%%%%%%%%%%%%%%%%%%%%%%%%%%%%%%%%%%%%%%%%%
Next we aim to provide answers to the following questions. First, \emph{how do we generate meta-training data that allows to approximate SI?} Second, given that most architectures are trained with a limited sequence-length, \emph{how does this affect the meta-training protocol of neural models?} Third, \emph{can we use different program distributions (making interesting programs more likely) without losing universality?} 

%-------------------------------------------------------------%
\subsection{The right dataset: Estimating Solomonoff from Solomonoff Samples} \label{sec:solomonoff_samples}
%-------------------------------------------------------------%
Our aim here is to define a data generation process such that we obtain an approximation to $M$ (see Figure~\ref{fig:summary}) when training our model $\pi_\theta$ on it (assuming for now universality and essentially infinite capacity). We consider the incomputable and computable cases. All proofs can be found in the Appendix~\ref{sec:samples_details}.

\textbf{Solomonoff Data Generator (incomputable).\,}
Putting uniform random bits $p$ on the (read-only) input tape of a monotone UTM $U$ generates a certain distribution $M$ of (in)finite strings $x$ on the output tape.
This is exactly Solomonoff's prior $M$ and a semimeasure (see Section~\ref{sec:background}).
Sampling from $M$ is trivial; we just described how and coincides exactly with the standard meta-learning setup where programs correspond to tasks.
$M$ is equivalent to the more formal Definition~\ref{def:monotone_solomonoff_prior}. The following proposition shows consistency.

% \begin{proposition}
\begin{restatable}{proposition}{solinfinite}\label{prop:solinfinite}
Let $D:=(x^1,...,x^J)$ be $J$ (in)finite sequences sampled from a semimeasure $\mu$ (e.g. $M$).
We can estimate $\mu$ as follows:
$
\hat\mu_D(x) ~:=~ \frac1{|D|}\sum_{y\in D}[\![\ell(y)\geq\ell(x)~\wedge~y_{1:\ell(x)}=x]\!]
  ~\stackrel{w.p.1}\longrightarrow \mu(x) ~~\text{for}~~ |D|\to\infty$.
% \begin{align*} %\label{eq:smllm}
%   \hat\mu_D(x) ~:=~ \frac1{|D|}\sum_{y\in D}[\![\ell(y)\geq\ell(x)~\wedge~y_{1:\ell(x)}=x]\!]
%   ~~~\stackrel{w.p.1}\longrightarrow~~~ \mu(x) ~~~\text{for}~~~ |D|\to\infty
% \end{align*}
\end{restatable}
% \end{proposition}

 Unfortunately there are three infinities which prevent us from using $M$ above.
There are infinitely many programs, 
programs may loop forever, 
and output strings can have infinite length.  Therefore, we define the following computable version of the Solomonoff prior.
\begin{definition}[Computable Solomonoff Prior]\label{def:computable_solomonoff}
Let programs be of length $\leq L$ and  stop $U$ after $s$ steps (denoted $U^s$), 
or if the output reaches length $n$. Then,
\ifICML
\begin{align*}
  M_{s,L,n}(x) := \sum_{p\in\{0,1\}^{\leq L}:U^s(p)=x*\hspace{-9ex}} 2^{-\ell(p)} \qquad  \begin{array}{l}
  \text{if}~~ \ell(x)\leq n \\
  \text{and} ~~0~~ \text{otherwise}\end{array}
\end{align*}
\else
\begin{align*}
  M_{s,L,n}(x) ~:=~ \sum_{p\in\{0,1\}^{\leq L}:U^s(p)=x*\hspace{-9ex}} 2^{-\ell(p)} ~~~\text{if}~~~ \ell(x)\leq n ~~~\text{and} ~~~0~~~ \text{otherwise}
\end{align*}
\fi
is a computable version of the Solomonoff prior and a semimeasure.
\end{definition}

We can sample $D^J:=(x^1,...,x^J)$ from $M_{s,L,n}$ in the same trivial way as described above for $M$, but now the involved computation is finite.
Note that all sampled strings have length $\leq n$, since $M_{s,L,n}(x):=0$ for $\ell(x)>n$. Consistency of meta-training data is shown next.

\begin{restatable}{proposition}{computablesolomonoff}\label{prop:computablesolomonoff}
% \begin{proposition}
Let  now $D^J:=(x^1,...,x^J)$ be samples from the measure $M_{s,L,n}$. Then,
% \begin{align*}
  $\hat M_{D^J}(x) = \frac1J\sum_{y\in D^J}[\![\ell(y)\geq\ell(x)~\wedge~y_{1:\ell(x)}=x]\!]
  ~~~\longrightarrow~~~ M_{s,L,n}(x) ~~~\text{for}~~~ J\to\infty
$.
% \end{align*}
% \end{proposition}
\end{restatable}

Since $M(x)=\lim_{s,L,n\to\infty} M_{s,L,n}(x)=\sup_{s,L,n} M_{s,L,n}(x)$,
we in particular have $\hat M_{D^J}\rightarrow M$ for $s,L,n,J\to\infty$.
Note that $D^J$ depends on $s,L,n$, but this can easily be avoided 
by choosing $s(j),L(j),n(j)$ to be any functions tending to infinity,
and sampling $x^j$ from $M_{s(j),L(j),n(j)}(x)$ for $j=1,2,3,...$.

\begin{remark}\label{rem:inconveniences_semimeasure}
  Although $M_{s,L,n}$ is computable, it still suffers from two inconveniences. First, sampling from it is inefficient because it is a semimeasure and exhibits a probability gap. Second, we need to differentiate whether programs halt or end up in a infinite non-printing loop (to fill the probability gap with ``absorbing'' tokens when training).
   We can bypass these inconveniences by estimating the normalized and computable Solomonoff prior combining Definitions~\ref{def:normalized_solomonoff_prior} and~\ref{def:computable_solomonoff}. 
\end{remark}

We can estimate the (computable) normalized Solomonoff prior, $M_{s,L,n}^{norm}(x)$, by the following.
\begin{restatable}{proposition}{normalizedconvergence}\label{prop:normalized_convergence}
Using the definitions from Proposition~\ref{prop:computablesolomonoff} we have that
\ifICML
\begin{align*}
  \hat M_{s,L,n}^{norm}(x_t|x_{<t}) & ~=~  
  \frac{\sum_{y\in D^J}[\![\ell(y)\geq t~\wedge~y_{1:t}=x_{1:t}]\!]}{\sum_{y\in D^J}[\![\ell(y)\geq t~\wedge~y_{<t}=x_{<t}]\!]} \\
  &~~~\stackrel{J\to\infty}\longrightarrow~~~ M_{s,L,n}^{norm}(x_t|x_{<t})
\end{align*}
\else
\begin{align*}
  \hat M_{s,L,n}^{norm}(x_t|x_{<t}) ~=~ 
  \frac{\sum_{y\in D^J}[\![\ell(y)\geq t~\wedge~y_{1:t}=x_{1:t}]\!]}{\sum_{y\in D^J}[\![\ell(y)\geq t~\wedge~y_{<t}=x_{<t}]\!]}
  ~~~\stackrel{J\to\infty}\longrightarrow~~~ M_{s,L,n}^{norm}(x_t|x_{<t})
\end{align*}
\fi
Then, we can take the product over $t=1,...,n$ to obtain $\hat M_{s,L,n}^{norm}(x)\to M_{s,L,n}^{norm}(x)\to M^{norm}(x)$. 
\end{restatable}

\textbf{Summary.} Propositions~\ref{prop:solinfinite}, ~\ref{prop:computablesolomonoff} and ~\ref{prop:normalized_convergence} state that the data generated by the Solomonoff Data Generator and their respective variants (computable and normalized computable) are statistically consistent, and that meta-training on this data would make an estimator converge to their respective Solomonoff version (under realizability and learnability assumptions).

%-------------------------------------------------------------%
\subsection{Training Models on Solomonoff Data using Fixed-Sequence Lengths} \label{sec:LLMforM} % 2309(1c)
%-------------------------------------------------------------%

Most neural models (especially Transformers) require training sequences of fixed length $n$. Due to this, we require a slight modifications to the loss function for shorter-than-$n$ sequences to maintain convergence to SI. 
We drop $s,L,n$ from $M_{s,L,n}^{\cdots}$ since what follows holds for infinite as well as finite values. We focus on describing the training protocol that converges to the normalized version of Solomonoff, $M^{norm}$. We refer readers interested in the standard unnormalized version ($M$) to the Appendix~\ref{sec:training_llm_details}.

\textbf{Normalized Solomonoff $M^{norm}$ with neural networks.} 
To converge to $M^{norm}$, we pad the $x^j$ in $D^J$ to length $n$ with arbitrary symbols from $\cX$, and cut the log-loss short at $\ell(x^j)$. When doing so, the log-loss takes the form (see Appendix~\ref{sec:normalized_solomonoff_loss} for derivation that uses Proposition~\ref{prop:normalized_convergence}):
\ifICML
\begin{align}\label{eq:loss_normalized_solomonoff}
   \text{Loss}(\theta)~=~ -\sum_{t=1}^n\sum_{x_{<t}}\bigg( \Big(\sum_{x_t}\hat M_{D^J}(x_{1:t})\Big) \qquad \\
  \qquad \Big(\sum_{x_t}\hat M^{norm}(x_t|x_{<t})\log \pi_\theta(x_t|x_{<t})\Big) \bigg)\nonumber
\end{align}
\else
\begin{equation}\label{eq:loss_normalized_solomonoff}
   \text{Loss}(\theta)~=~ -\sum_{t=1}^n\sum_{x_{<t}}\Big(\sum_{x_t}\hat M_{D^J}(x_{1:t})\Big)\Big(\sum_{x_t}\hat M^{norm}(x_t|x_{<t})\log \pi_\theta(x_t|x_{<t})\Big) 
\end{equation}
\fi
In this form, it is easy to see how the last bracket, and hence the loss, is minimized for $\pi_\theta(x_t|x_{<t})=\hat M^{norm}(x_t|x_{<t})$, as desired.
By the chain rule this implies that the neural model $\pi_\theta(x)$ converges to $\hat M^{norm}(x)$.
Note that $\text{Loss}(\theta)$ does \emph{not} depend on the padding of $x^j$, so any padding leads to the same gradient and same solution.

Under the (unrealistic) assumptions that the neural model has the capacity to represent $\hat M^{\cdots}$,
and the learning algorithm can find the representation, 
this (tautologically) implies that the neural model distribution $\pi_\theta$ converges to $\hat\mu=\hat M^{\cdots}$.
Similarly, if the neural model is trained on $x^j$ sampled from $M_{s(j),L(j),n}^{\cdots}(x)$ for $j=1,2,3,...$,
it converges to $M_{\infty,\infty,n}^{\cdots}$. 
For a neural model with context length $n$ increasing over time,
even $\hat M^{\cdots}\to M^{\cdots}_{\infty,\infty,\infty}$ could be possible. Though theoretically possible, there are many practical challenges that need to be surmounted to achieve this, one of them being how to efficiently sample programs.

%-------------------------------------------------------------%
\subsection{Solomonoff from Non-Uniform Samples}\label{sec:SolNU} % 2309(3)
%-------------------------------------------------------------%

For practical purposes, sampling from non-uniform (possibly learned) distribution over programs can be advantageous for efficiency. For our BrainPhoque language (that we use in our experiments later) it increases the yield of `interesting' programs by a factor of 137 (see Appendix Table~\ref{tab:mcsampling}). Below we show this can be done without any concerns on losing universality.

Let $Q$ be a probability measure on $\cX^\infty$,
with shorthand $Q(q):=Q(\Gamma_q)$, the $Q$-probability that a sequence starts with $q$,
where $\Gamma_q:=\{\omega\in\cX^\infty:q\sqsubseteq\omega\}=q\cX^\infty$.
We define the \emph{generalized Solomonoff semimeasure} as
\ifICML
\begin{align*} %\label{eq:gsol}
  & M_T^Q(x) ~:= \sum_{q:T(q)=x*\!\!\!\!}Q(q)
%   & ~~~\text{with special case}~~~
%   M_U(x) ~:= \sum_{q:U(q)=x*\!\!\!\!}2^{-\ell(q)}
\end{align*}
  with special case $M_U(x) ~:= \sum_{q:U(q)=x*}2^{-\ell(q)}$
\else
\begin{align*} %\label{eq:gsol}
  M_T^Q(x) ~:= \sum_{q:T(q)=x*\!\!\!\!}Q(q) ~~~\text{with special case}~~~
  M_U(x) ~:= \sum_{q:U(q)=x*\!\!\!\!}2^{-\ell(q)}
\end{align*}
\fi
for a universal TM $T=U$ and unbiased coin flips $Q(q)=2^{-\ell(q)}$.
$M_U$ is strongly universal in the sense that it is a Bayesian mixture over all lower semi-computable semimeasures \citep{Hutter:11unipreq}. Next, we show that under very mild conditions on $Q$, $M_U^Q$ is also universal. This finding is similar to~\citep{sterkenburg2017generalized}, but our independently discovered proof is shorter and more self-contained. 
\begin{restatable}[Universality of generalized Solomonoff semimeasures]{theorem}{ugsol}\label{thm:ugsol}
$M_U^Q(x)$ is strongly universal, provided $Q$ is a computable measure and $Q(q)>0~\forall q\in\cX^*$ and $Q(q_{1:n})\to 0$ for $n\to\infty$.
More precisely, for all universal monotone TM $U$ and all $Q$ with the above properties, 
there exists a universal MTM $V$ (as constructed in the proof) s.th.\ $M_U^Q(x)=M_V(x)~\forall x$. Proof in Appendix~\ref{sec:generalized_solomonoff_proof}.
\end{restatable}

\textbf{Note on the assumptions above.} 
We assumed an infinite number of data points and universality (and learnablity) of the approximator, which are difficult to obtain in practice and diminish the relevance of inductive biases of neural models. For finite data, however, inductive biases are important for strong generalization. We leave out of the scope of the paper the theoretical work on the effect of the inductive bias and universality of neural models and simply provide experimental evidence of neural network performance in the next section.

%%%%%%%%%%%%%%%%%%%%%%%%%%%%%%%%%%%%%%%%%%%%%%%%%%%%%%%%%%%%%%%
\section{Experimental Methodology}\label{sec:methods}
%%%%%%%%%%%%%%%%%%%%%%%%%%%%%%%%%%%%%%%%%%%%%%%%%%%%%%%%%%%%%%%
\begin{figure*}[t!]
% \centering
     \begin{subfigure}[b]{0.40\textwidth}
        \centering
        \includegraphics[width=\textwidth]{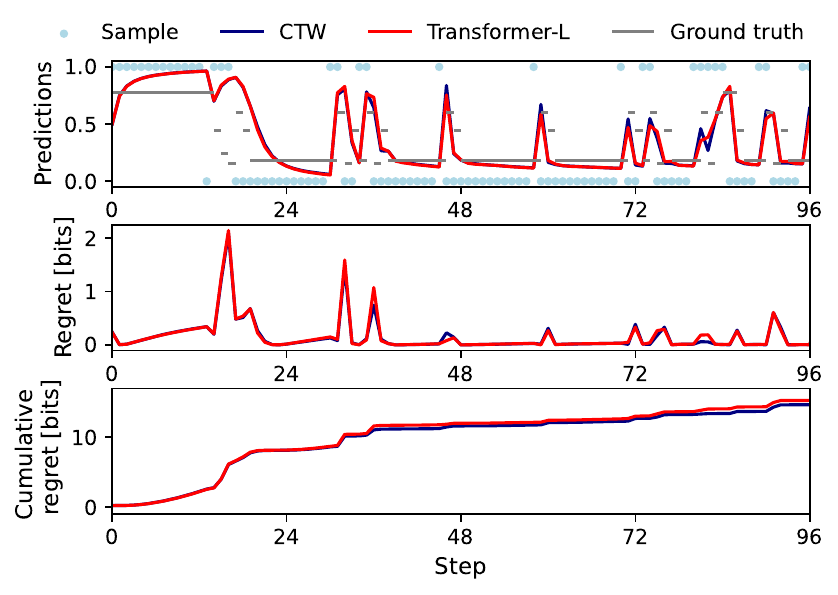}
        % \caption{Single sequence example.}
    \end{subfigure}
    % \hspace{1pt}
    \hfill
    \begin{subfigure}[b]{0.32\textwidth}
        \centering
        \includegraphics[width=\textwidth]{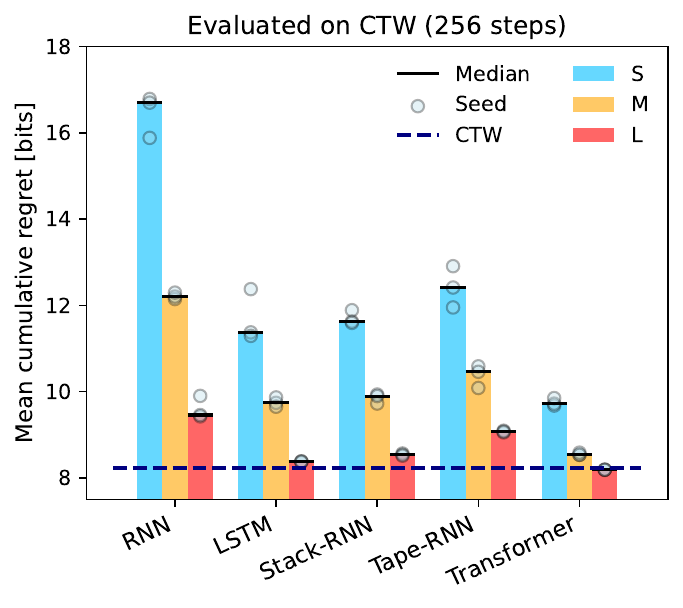}
        % \caption{Evaluation on $6$k sequences.}
        % \label{fig:ctw_redundancies}
    \end{subfigure}
    \hfill
    \begin{subfigure}[b]{0.23\textwidth}
        \includegraphics[width=\textwidth, trim={454, 0, 0, 0}, clip]{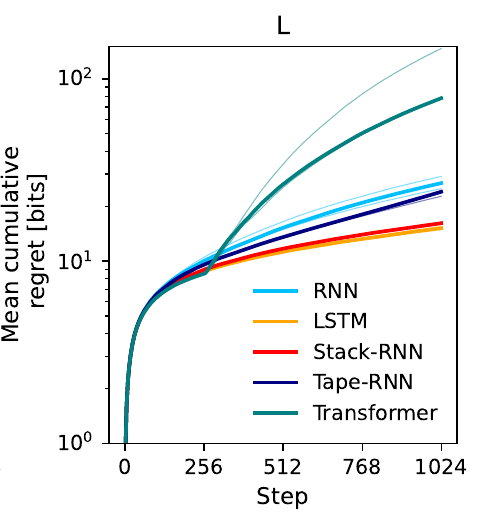}
        \vspace{2pt}
    \end{subfigure}
    \caption{Evaluation on VOMS data. \textbf{Left:} Example sequence and highly overlapped predictions of Transformer-L (red) and Bayes-optimal CTW predictor (blue). Lower panels show instantaneous and cumulative regret w.r.t. the ground-truth. \textbf{Middle:} Mean cumulative regret over $6$k sequences (length~$256$, max. CTW tree depth~$24$, in-distribution) for different networks ($3$ seeds) and sizes (S, M, L). Larger models perform better for all architectures, and the Transformer-L and LSTM-L match the optimal CTW predictor. \textbf{Right:} Length generalization ($1024$ steps). LSTMs generalize to longer length, whereas Transformers do not. \label{fig:ctw_results}}
        
\end{figure*}

We aim to evaluate various neural architectures and sizes trained on UTM and two other types of algorithmically generated data for comparison and analysis. 

\textbf{Variable-order Markov Sources (VOMS).}
A $k$-Markov model assigns probabilities to a string of characters by, at any step $t$, only using the last $k$ characters to output the next character probabilities. A VOMS is a Markov model where the value of $k$ is variable and it is obtained using a tree of non-uniform depth. A tree here is equivalent to a program that generates data. We sample trees and meta-train on the generated data. We consider \emph{binary} VOMS where a Bayes-optimal predictor exists: the Context Tree Weighting (CTW) predictor~\citep{willems1995context,willems1997reflections}, to which we compare our models to. CTW is only universal w.r.t. $n$-Markov sources, and not w.r.t. all computable functions  like SI. See Appendix~\ref{sec:ctw_details} for more intuition on VOMS, how we generate the data and how to compute the CTW Bayes-optimal predictor.

\textbf{Chomsky Hierarchy (CH) Tasks.}  We take the $15$ algorithmic tasks (e.g. arithmetic, reversing strings) from~\cite{deletang2022neural} lying on different levels of the Chomsky hierarchy (see Appendix~\ref{sec:chomsky_method_details} for a description of all tasks). These tasks are useful for comparison and for assessing the algorithmic power of our models. In contrast to~\cite{deletang2022neural}, in which they train on \emph{individual} tasks, we are interested in meta-training on all tasks \emph{simultaneously}. 
We make sure that  all tasks use the same alphabet $\mathcal X$ (expanding the alphabet of tasks with smaller alphabets).  We do not consider transduction as in~\cite{deletang2022neural} but sequence prediction, thus we concatenate inputs and outputs with additional delimiter tokens i.e. for $\{(x_i \in \mathcal X, y_i \in \mathcal X)\}_{i=1}^I$ and delimiters `$,$' and `$;$', we construct sequences of the form $z:=(x_1,y_1;x_2,y_2; \dots x_n,y_n; \dots )$. We evaluate our models using the regret (and accuracy) \emph{only} on the output symbols, masking the inputs because they are usually random and non-informative of task performance.  Denoting $\mathcal O_z$ the set of outputs time-indices, we compute accuracy for trajectory $z$ as $A(z) :=  \frac{1}{| \mathcal O_z |}\sum_{t \in \mathcal O_z} [\![ \argmax_y \pi_\theta (y | z_{<t}) = z_t  ]\!]$.
See Appendix~\ref{sec:chomsky_method_details} for details.

\textbf{Universal Turing Machine Data.} Following Sections~\ref{sec:solomonoff_samples} and~\ref{sec:LLMforM}, we generate random programs (encoding any structured sequence generation process) and run them in our UTM to generate the outputs.  A program could, in principle, generate the image of a cow, a chess program, or the books of Shakespeare, but of course, these programs are extremely unlikely to be sampled (see Figure~\ref{fig:brainphoque} in the Appendix for exemplary outputs). 
As a choice of UTM, we constructed a variant of the BrainF*ck UTM~\citep{brainfck}, which we call BrainPhoque,
mainly to help with the sampling process and to ensure that all sampled programs are valid. We set output symbols alphabet size to $|\mathcal X|=17$, equal to the Chomsky tasks, to enable task-transfer evaluation.
BrainPhoque has a single working tape and a write-only output tape.
It has $7$ instructions to move the working tape pointer (WTP),
de/increment the value under the WTP (the \emph{datum}), perform jumps and append the datum to the output.
We skip imbalanced brackets to make all programs valid. While it slightly changes the program distribution, this is not an issue according to Theorem~\ref{thm:ugsol}: each valid program has a non-zero probability to be sampled.
Programs are generated and run at the same time, as described in Sections~\ref{sec:solomonoff_samples} and~\ref{sec:LLMforM}, for $s=1000$ steps with $200$ memory cells, with a maximum output length of $n=256$ symbols.
Ideally, we should use SI as the optimal baseline comparison but since it is uncomputable and intractable, we calculate a (rather loose, but non-trivial) upper bound on the log-loss by using the prior probability of shortened programs (removing unnecessary brackets or self-canceling instructions) that generate the outputs. See Appendix~\ref{sec:utm_methods_details} for a full description of BrainPhoque and our sampling procedure.

\textbf{Neural Predictors.} Our neural models $\pi_\theta$ sequentially observe symbols $x_{<t}$ from the data generating source and predict the next-symbol probabilities $\pi_\theta (\cdot | x_{<t})$. We train our models using the log-loss $\text{Loss}(\theta) := -\frac{1}{n} \sum_{t=1}^n \log \pi_\theta (x_t | x_{<t})$, therefore maximizing lossless compression of input sequences~\citep{deletang2023compress}. We use stochastic gradient descent with the ADAM optimizer~\citep{kingma2014adam}. We train for $500$K iterations with batch size $128$, sequence length $256$, and learning rate $10^{-4}$. On the UTM data source, we cut the log-loss to approximate the normalized version of SI (see Section~\ref{sec:LLMforM}). We evaluate the following architectures: RNNs, LSTMs, Stack-RNNs, Tape-RNNs and Transformers. We note   that Stack-RNNs~\citep{joulin2015inferring} and Tape-RNNs~\citep{deletang2022neural} are RNNs augmented with a stack and tape memory, respectively, which stores and manipulate symbols. This external memory should help networks to predict better, as showed in~\citet{deletang2022neural}. We consider three model sizes (S, M and L) for each architecture by increasing the width and depth simultaneously. We train $3$ parameter initialization seeds per model variation. See Appendix~\ref{sec:architecture_details} for all architecture details.

\textbf{Evaluation procedure.} Our main evaluation metric is  the \emph{expected instantaneous regret}, $R_{\pi\mu}(t) := \mathbb E_{x_t \sim \mu} \left[ \log \mu(x_t\mid x_{<t}) - \log \pi (x_t\mid x_{<t}) \right]$ (at time $t$),  and \emph{cumulative expected regret}, $R_{\pi\mu}^T := \sum_{t=1}^T  R_{\pi\mu}(t)$, where $\pi$ is the model and  $\mu$ the ground-truth source. The lower the regret the better. We evaluate our neural models on $6$k sequences of length $256$, which we refer as \emph{in-distribution} (same length as used for training) and of length $1024$, referred as \emph{out-of-distribution}.

%%%%%%%%%%%%%%%%%%%%%%%%%%%%%%%%%%%%%%%%%%%%%%%%%%%%%%%%%%%%%%%
\section{Results}
%%%%%%%%%%%%%%%%%%%%%%%%%%%%%%%%%%%%%%%%%%%%%%%%%%%%%%%%%%%%%%%

%-------------------------------------------------------------%
% \subsection{Context Tree Weighting}
%-------------------------------------------------------------%

\begin{figure*}[t]
    % \centering
    \begin{subfigure}[b]{0.33\textwidth}
        \centering
        \includegraphics[width=\textwidth]{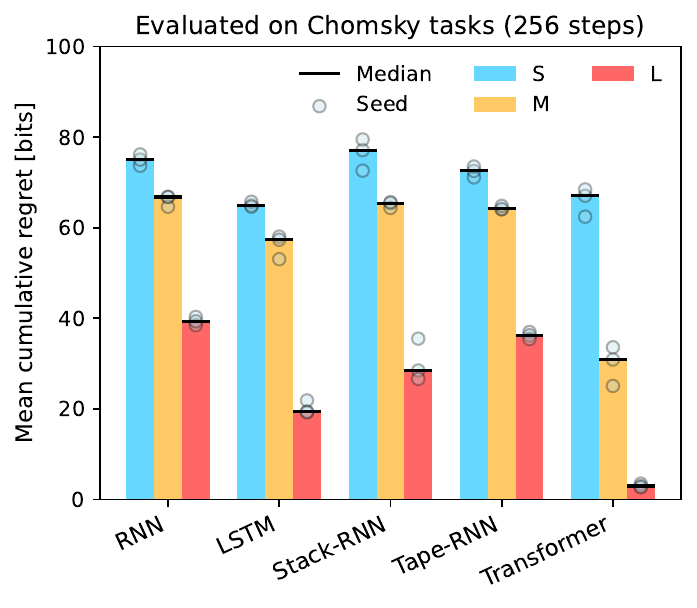}
        % \caption{Mean cumulative regret.}
    \end{subfigure}
    \hfill
    \hspace{1pt}
    \begin{subfigure}[b]{0.33\textwidth}
        \centering
        \includegraphics[width=\textwidth]{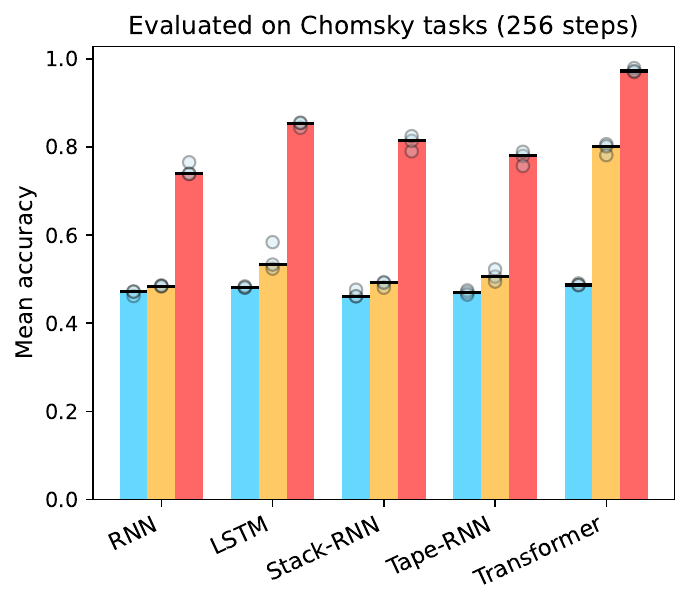}
        % \caption{Mean accuracy.}
    \end{subfigure}
    \hfill
    \begin{subfigure}[b]{0.28\textwidth}
        \includegraphics[width=0.9\textwidth, trim={458, 0, 0, 0}, clip]{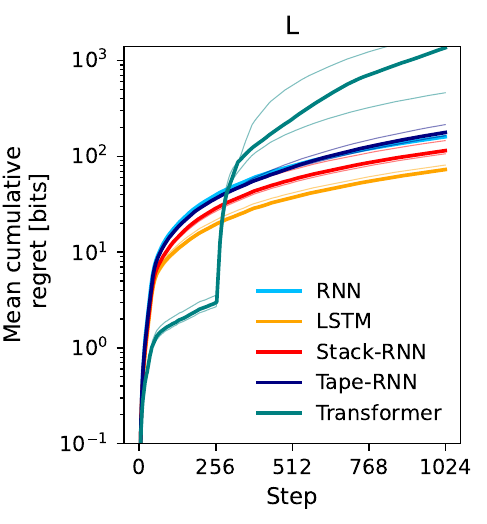}
        \vspace{5pt}
    \end{subfigure}
    \caption{Evaluation on $6$k sequences from the \textbf{Chomsky hierarchy tasks} ($400$ per task). As the model size increases, cumulative regret (\textbf{Left}) and accuracy (\textbf{Middle}) improve across all architectures. Overall, the Transformer-L achieves the best performance by a margin. \textbf{Right:} Length generalization ($1024$ steps).  Detailed results per task are in Figure~\ref{fig:chomsky_results_by_task} on the Appendix.}
    \label{fig:main_results_chomsky}
\end{figure*}

\textbf{Variable-order Markov Source (VOMS)  Results.} In Figure~\ref{fig:ctw_results} (Left) we show an example trajectory from VOMS data-source of length $256$ with the true samples (blue dots), ground truth (gray), Transformer-L (red) and CTW (blue) predictions. As we can see, the predictions of the CTW predictor and the Transformer-L are overlapping, suggesting that the Transformer is implementing a Bayesian mixture over programs/trees like the CTW does, which is necessary to perform SI. In the second and third panels the instantaneous regret and the cumulative regret also overlap. Figure~\ref{fig:ctw_results} (Middle) shows the cumulative regret of all neural predictors evaluated in-distribution.  First, we observe that as model size increases (from S, M, to L) the cumulative regret decreases. The best model is the Transformer-L achieving optimal performance, whereas the worst models are the RNNs and the Tape-RNNs. The latter model likely could not successfully leverage its external memory. Note how LSTM-L achieves close to optimal performance. On the Right we show the out-of-distribution performance showing how transformers fail on length-generalization, whereas LSTMs perform the best. To better understand where our models struggle, we show in the Appendix~\ref{sec:ctw_result_details}, Figures~\ref{fig:redundancies_vs_tree_depth} and~\ref{fig:redundancies_vs_context_length},  the cumulative regret averaged across trajectories from different CTW tree depths and context lengths. Models perform uniformly for all tree-depths and struggle on mid-sized context-lengths.

%-------------------------------------------------------------%
% \subsection{Chomsky Hierarchy}
%-------------------------------------------------------------%

\textbf{Chomsky Hierarchy Results.} In Figure~\ref{fig:main_results_chomsky} (Left) we show the in-distribution performance of all our models trained on the Chomsky hierarchy tasks by means of cumulative regret and accuracy. Overall, the Transformer-L achieves the best performance by a margin. This suggests that our models, specially Transformers, have the capability of algorithmic reasoning to some extent. On  the Right we show the length-generalization capabilities of models, showing how Transformers fail to generalize to longer lengths. In the Appendix (Figure~\ref{fig:chomsky_results_by_task}) we show the results for each task individually.

%-------------------------------------------------------------%
% \subsection{Universal Turing Machines}
%-------------------------------------------------------------%

\begin{figure*}[t]
    % \centering
    \begin{subfigure}[b]{0.33\textwidth}
        \centering
        \includegraphics[width=\textwidth]{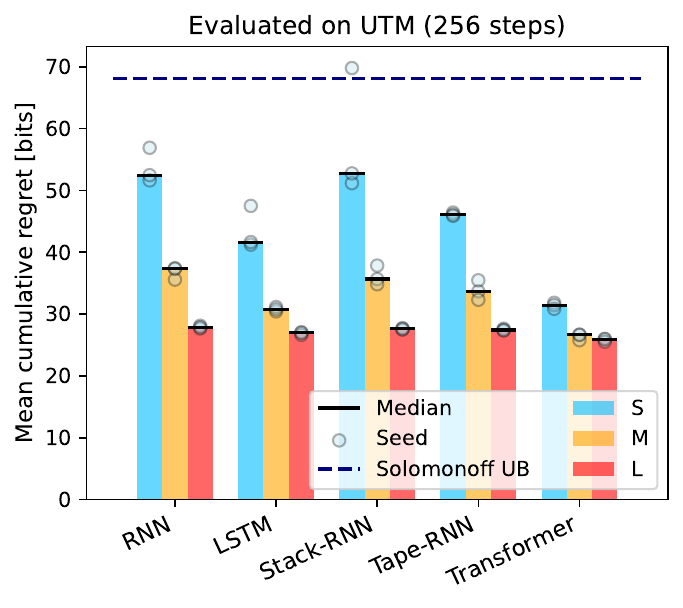}
        % \caption{Mean cumulative regret.}
    \end{subfigure}
    % \hspace{1pt}
    \hfill
    \begin{subfigure}[b]{0.33\textwidth}
        \centering
        \includegraphics[width=\textwidth]{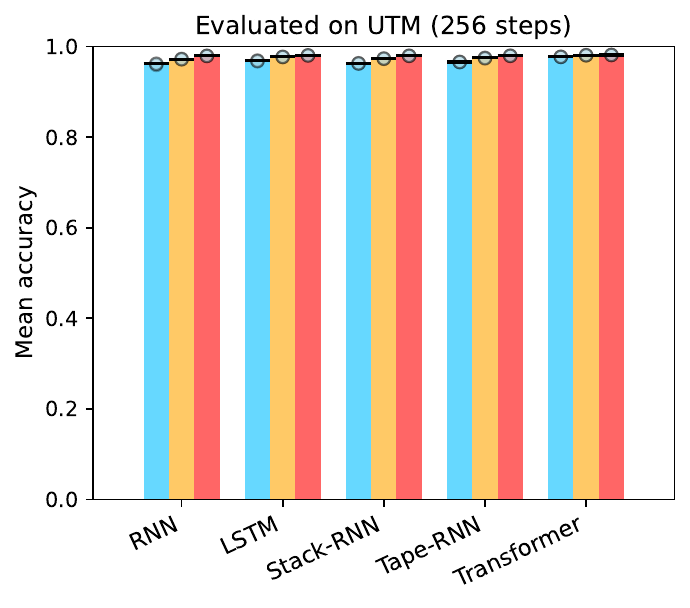}
        % \caption{Mean accuracy.}
    \end{subfigure}
    \hfill
    \begin{subfigure}[b]{0.28\textwidth}
        \includegraphics[width=0.9\textwidth, trim={454, 0, 0, 0}, clip]{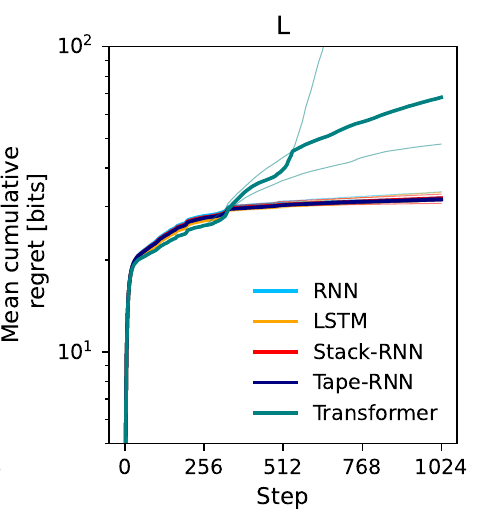}
        \vspace{5pt}
    \end{subfigure}
    \caption{Evaluation on the \textbf{UTM data generator} with $6$k sequences. \textbf{Left:} The larger the architecture the lower the cumulative regret. We see better performance than the non-trivial baseline Solomonoff Upper Bound (UB).  \textbf{Middle:} The mean accuracy on UTM data shows the models can quickly learn UTM patterns. \textbf{Right:} Length generalization ($1024$ steps). Detailed results per program length are in Figure~\ref{fig:results_per_program_length_utm}.}
    \label{fig:main_results_utm}
\end{figure*}

\begin{figure*}[t]
\centering
\begin{subfigure}[b]{0.3\textwidth}
    \includegraphics[width=\textwidth]{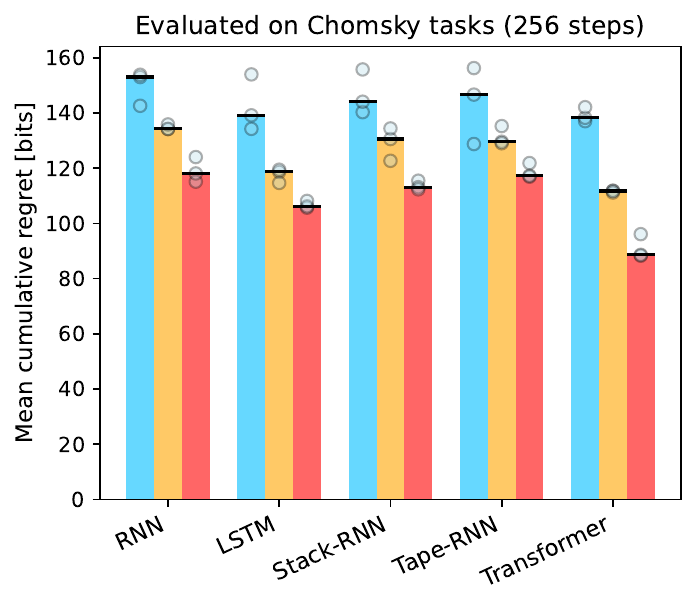}
    % \caption{Regret}
\end{subfigure}
\begin{subfigure}[b]{0.3\textwidth}
    \includegraphics[width=\textwidth]{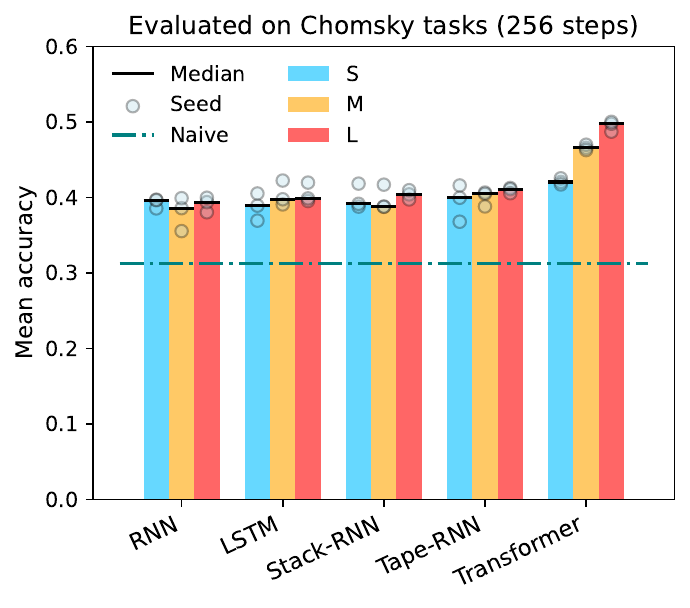}
    % \caption{Accuracy}
\end{subfigure}
\hfill
\begin{subfigure}[b]{0.189\textwidth}
    \includegraphics[width=\textwidth]{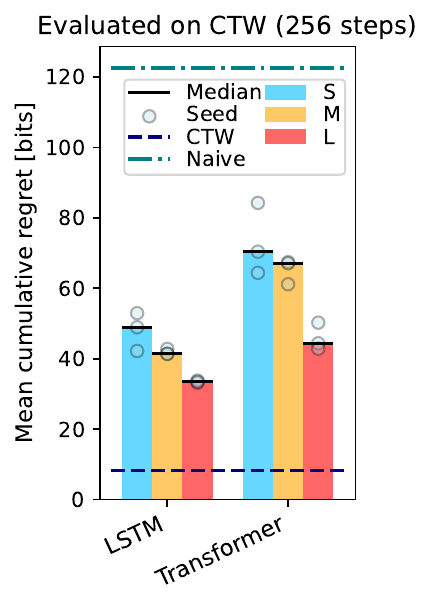}
    % \caption{Accuracy}
\end{subfigure}
\begin{subfigure}[b]{0.186\textwidth}
    \includegraphics[width=\textwidth]{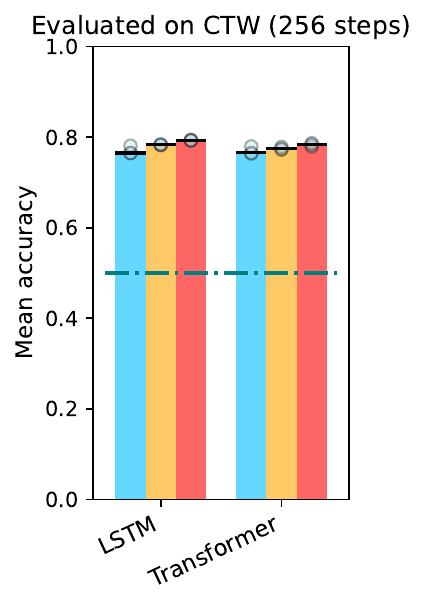}
    % \caption{Accuracy}
\end{subfigure}
    \caption{\textbf{Transfer learning} from \emph{UTM-trained models} on $3$k trajectories. Mean cumulative regret (\textbf{Left}) and accuracy (\textbf{Middle-Left}) of neural models trained on UTM data evaluated against the tasks of the Chosmky hierarchy. We observe a small increase in accuracy (transfer) from the Transformer models. Transfer to CTW is shown in the right two panels: \textbf{Middle-Right:} mean cumulative regret, \textbf{Right:} mean accuracy; `Naive' is a random uniform predictor.}
     \label{fig:utm_to_chomsky_transfer}
\end{figure*}

\textbf{Universal Turing Machine Results.} Figure~\ref{fig:main_results_utm} (Left) shows the mean cumulative regret on the UTM task with the (loose) Solomonoff Upper Bound (UB) as a non-trivial baseline (see Section~\ref{sec:methods} for its description). In the Middle we show how all models achieve fairly good accuracy. This shows how our models are capable of learning a broad set of patterns present in the data (see example UTM trajectories in appendix Figure~\ref{fig:brainphoque}). In general, larger architectures attain lower cumulative regret and all models beat the Solomonoff upper bound. Performing better than the bound is non-trivial  since the upper-bound is computed using the underlying program that generated the outputs whereas the neural models do not have this information. In Figure~\ref{fig:results_per_program_length_utm} (in the Appendix) we show the cumulative regret against program length and, as expected, observe that the longer the underlying program of a sequence the higher the cumulative regret of our models, suggesting a strong correlation between program length and prediction difficulty.
Remarkably, in Figure~\ref{fig:utm_to_chomsky_transfer} we see that the Transformer networks trained on UTM data exhibit the most transfer to the Chomsky tasks and, LSTMs transfer the most to the VOMS task (compare to the `naive' random predictor).  For the VOMS, we re-trained the LSTM and Transformer models with the BrainPhoque UTM setting the alphabet size to $2$ matching our VOMS task to enable comparison. All transfer results suggest that UTM data contains enough transferable patterns for these tasks.

%%%%%%%%%%%%%%%%%%%%%%%%%%%%%%%%%%%%%%%%%%%%%%%%%%%%%%%%%%%%%%%
\section{Discussion and Conclusions}
%%%%%%%%%%%%%%%%%%%%%%%%%%%%%%%%%%%%%%%%%%%%%%%%%%%%%%%%%%%%%%%

\textbf{Large Language Models (LLMs) and Solomonoff Induction.} The last few years the ML community has witnessed the training of enormous models on massive quantities of diverse data~\citep{kenton2019bert,hoffmann2022training}. This trend is in line with the premise of our paper, i.e. to achieve increasingly universal models one needs large architectures and large quantities of diverse data. 
LLMs have been shown to have impressive in-context learning capabilities~\citep{kenton2019bert,chowdhery2022palm}. 
LLMs pretrained on long-range coherent documents can learn new tasks from a few examples by inferring a shared latent concept~\citep{xie2022an,wang2023large}. They can do so because in-context learning does implicit Bayesian inference  (in line with our CTW experiments) and builds world representations and algorithms~\citep{li2023emergent,li2023transformers} (necessary to perform SI). 
In fact, one could argue that the impressive in-context generalization capabilities of LLMs is a sign of a rough approximation of Solomonoff induction. 
The advantage of pre-trained LLMs compared to our method (training on universal data) is that LLM data (books, code, online conversations etc.) is generated by humans, and thus very well aligned with the tasks we (humans) want to solve; whereas our UTMs do not necessarily assign high probability to human tasks. 

\textbf{Learning the UTM.} Theorem~\ref{thm:ugsol} of our paper (and~\citep{sterkenburg2017generalized}) opens the path for modifying/learning the program distribution of a UTM while maintaining the universality property. This is of practical importance since we would prefer distributions that assign high probability to programs relevant for human tasks. Similarly, the aim of~\cite{sunehag14learnutm} is to directly learn a UTM aligned to problems of interest. A good UTM or program distribution would contribute to having better synthetic data generation used to improve our models. This would be equivalent to data-augmentation technique so successfully used in the machine learning field~\citep{perez2017effectiveness,lemley2017smart,kataoka2020pre}. In future work, equipped with our Theorem~\ref{thm:ugsol}, we plan study optimizations to the sampling process from UTMs to produce more human-aligned outputs.

\textbf{Increasingly Universal Architectures.}
The output of the UTM $U^s(p)$ (using program $p$) requires at maximum $s$ computational steps.  Approximating $M_{s, L, n}$ would naively require wide networks (to represent many programs in parallel) of $s$-depth and context length $n$. Thus bigger networks would better approximate stronger SI approximations. If computational patterns can be reused, depth could be smaller than $s$. Transformers seem to exhibit reusable ``shortcuts'' thereby representing all automata of length $T$ in $O(\log T)$-depth~\citep{liu2023transformers}.
An alternative way to increase the amount of serial computations is with chain-of-thought~\citep{wei2022chain} (see~\citet{hahn2023theory} for theoretical results).
When data is limited, inductive biases are important for generalization. Luckily it seems neural networks have an implicit inductive bias towards simple functions at initialization~\citep{dingle2018input,valle2018deep,mingard2023deep} compatible with Kolmogorov complexity, which is  greatly convenient when trying to approximate SI in the finite-data regime.

\textbf{Limitations.} Given the empirical nature of our results, we cannot guarantee that our neural networks mimic SI's universality. Solomonoff Induction is uncomputable/undecidable and one would need infinite time to exactly match it in the limit. However, our theoretical results establish that good approximations are obtainable, in principle, via meta-training; whereas our empirical results show that is possible to make practical progress in that direction, though many questions remain open, e.g., how to construct efficient relevant universal datasets for meta-learning, and how to obtain easily-trainable universal architectures.

\textbf{Conclusion.}  
We aimed at using meta-learning as driving force to approximate Solomonoff Induction. For this we had to carefully specify the data generation process and the training loss so that the convergence (to various versions of SI) is attained in the limit. Our experiments on the three different algorithmic data-sources tell that: neural models can implement algorithms and Bayesian mixtures, and that larger models attain increased performance. Remarkably, networks trained on the UTM data exhibit transfer to the other domains suggesting they learned a broad set of transferable patterns. We believe that we can improve future sequence models by scaling our approach using UTM data and mixing it with existing large datasets.

\newpage

\ifICML 
\textbf{Impact Statement.}
This paper presents work whose goal is to advance the field of Machine Learning. There are many potential societal consequences of our work, none which we feel must be specifically highlighted here.
\else
\textbf{Reproducibility Statement.} On the theory side, we wrote all proofs in the Appendix. For data generation, we fully described the variable-order Markov sources in the Appendix; we used the open-source repository \url{https://github.com/google-deepmind/neural_networks_chomsky_hierarchy} for the Chomsky tasks and fully described our UTM in the Appendix. We used the same architectures as~\citet{deletang2022neural} (which can be found in the same open-source repository) with modifications described in the Appendix. For training our models we used JAX \url{ https://github.com/google/jax}.
\fi

\bibliography{main}

%%%%%%%%%%%%%%%%%%%%%%%%%%%%%%%%%%%%%%%%%%%%%%%%%%%%%%%%%%%%%%%
\newpage

\ifICML
\else
\section{Appendix}
\fi
%%%%%%%%%%%%%%%%%%%%%%%%%%%%%%%%%%%%%%%%%%%%%%%%%%%%%%%%%%%%%%%
\appendix

\ifICML
\onecolumn
\fi

%%%%%%%%%%%%%%%%%%%%%%%%%%%%%%%%%%%%%%%%%%%%%%%%%%%%%%%%%%%%%%%
\section{Solomonoff samples}\label{sec:samples_details}
%%%%%%%%%%%%%%%%%%%%%%%%%%%%%%%%%%%%%%%%%%%%%%%%%%%%%%%%%%%%%%%

\paragraph{Sampling from semimeasures.}
We can sample strings from a semimeasure $\mu$ as follows:
Start with the empty string $x=\epsilon$. \\
With probability $\mu(a|x):=\mu(xa)/\mu(x)$ extend $x\leftarrow xa$ for $a\in\cX$. Repeat. \\
With probability $1-\sum_{a\in\cX}\mu(a|x)$ return $x$.

Let $D:=(x^1,...,x^J)$ be $J$ (in)finite sequences sampled from $\mu$.
If we only have these samples, we can estimate $\mu$ as follows:
\begin{align} %\label{eq:smllm}
  \hat\mu_D(x) ~:=~ \frac1{|D|}\sum_{y\in D}[\![\ell(y)\geq\ell(x)~\wedge~y_{1:\ell(x)}=x]\!]
  ~~~\stackrel{w.p.1}\longrightarrow~~~ \mu(x) ~~~\text{for}~~~ |D|\to\infty
\end{align}
\emph{Proof:} Let $D_x:=(y\in D:\ell(y)\geq\ell(x)~\wedge~y_{1:\ell(y)}=x)$ be the elements in $D$ that start with $x$.
Since $x^j$ are sampled i.i.d.\ from $\mu$, the law of large numbers implies
$|D_x|/|D|\to\mu(x)$ for $J\to\infty$.\qed

\paragraph{Limit normalization.}
A simple way of normalization is
\begin{align*}
  \widetilde M_{s,L,n}(x_{1:t}) ~:=~ \frac{\sum_{x_{t+1:n}} M_{s,L,n}(x_{1:n})}{\sum_{x_{1:n}}M_{s,L,n}(x_{1:n})}
  ~~~\text{for}~~~ t\leq n ~~~\text{and}~~~ 0 ~~~\text{else}
\end{align*}
This is a proper measure for sequences up to length $n$.
Sampling from it is equivalent to sampling from $M_{s,L,n}$ but discarding all sequences shorter than $n$.
Let $\widetilde D:=(x^j\in D^J:\ell(x^j)\geq n)$. Then
\begin{align*}
  \hat{\widetilde M}_{\widetilde D}(x) ~=~ \frac1{|\widetilde D|}\sum_{y\in\widetilde D}[\![y_{1:\ell(x)}=x]\!]
  ~~~\longrightarrow~~~ M(x) ~~~\text{for}~~~ s,L,n,J\to\infty  
\end{align*}
\emph{Proof:} First, $|\widetilde D|/|D|$ is the relative fraction of sequences that have length $n$,
and $\sum_{x_{1:n}} M_{s,L,n}(x_{1:n})$ is the probability that a sequence has length $n$, hence the former converges to the latter for $J\to\infty$.
Second, 
\begin{align*}
  \hat{\widetilde M}_{\widetilde D}(x_{1:n}) 
  ~&=~ \frac1{|\widetilde D|}\sum_{y\in\widetilde D}[\![y_{1:\ell(x)}=x_{1:n}]\!] 
  ~=~ \frac{|D|}{|\widetilde D|}\frac1{|D|}\sum_{y\in D}[\![\ell(y)\geq n~\wedge~ y_{1:\ell(x)}=x_{1:n}]\!] \\
  ~&=~ \frac{|D|}{|\widetilde D|}\hat M_{D^J}(x_{1:n}) 
  ~~~\stackrel{J\to\infty}\longrightarrow~~~ \frac{M_{s,L,n}(x_{1:n})}{\sum_{x_{1:n}}M_{s,L,n}(x_{1:n})} ~=~ \widetilde M_{s,L,n}(x_{1:n})
\end{align*}
Third, take the sum $\sum_{x_{t+1:n}}$ on both sides, and finally the limit $s,L,n\to\infty$ and set $x=x_{1:t}$. \qed

A disadvantage of this normalization scheme is that the probability of a sequence $x$ depends on $n$ even if $\ell(x)<n$,
while $M_{s,L,n}(x)$ and $M_{\cdots}^{norm}(x)$ below are essentially independent of $n$.

\solinfinite*

\emph{Proof:} Let $D_x:=(y\in D:\ell(y)\geq\ell(x)~\wedge~y_{1:\ell(y)}=x)$ be the elements in $D$ that start with $x$.
Since $x^j$ are sampled i.i.d.\ from $\mu$, the law of large numbers implies
$|D_x|/|D|\to\mu(x)$ for $J\to\infty$.\qed

\computablesolomonoff*
\emph{Proof:} It follows directly from Proposition~\ref{prop:solinfinite}.

\normalizedconvergence*

\emph{Proof:} For $x=x_{<t}$ and $a=x_t$, we have
\begin{align}
  \sum_{a\in\cX}\hat M_{D^J}(xa) ~&=~ \frac1J\sum_a\sum_{y\in D^J}[\![\ell(y)\geq\ell(xa)~\wedge~y_{1:\ell(xa)}=xa]\!] \nonumber \\ 
  ~&=~ \frac1J\sum_{y\in D^J}[\![\ell(y)\geq t~\wedge~\exists a:y_{1:t}=xa]\!] \nonumber \\ \label{eq:solnconv}
  \text{hence}~~&~ \hat M_{s,L,n}^{norm}(a|x) ~=~ \frac{\hat M_{D^J}(xa)}{\sum_a\hat M_{D^J}(xa)}
  ~~~\stackrel{J\to\infty}\longrightarrow~~~ \frac{M_{s,L,n}(ax)}{\sum_a M_{s,L,n}(ax)} ~=~ M_{s,L,n}^{norm}(a|x) 
\end{align}
\qed

%%%%%%%%%%%%%%%%%%%%%%%%%%%%%%%%%%%%%%%%%%%%%%%%%%%%%%%%%%%%%%%
\section{Training with Transformers}\label{sec:training_llm_details}
%%%%%%%%%%%%%%%%%%%%%%%%%%%%%%%%%%%%%%%%%%%%%%%%%%%%%%%%%%%%%%%

\paragraph{Using Transformers for estimating $M$.}
% unnormalized M
Most Transformer implementations require sequences of fixed length (say) $n$. 
We can mimic shorter sequences by introducing a special absorbing symbol $\bot\not\in\cX$,
and pad all sequences $x^j$ shorter than $n$ with $\bot$s.
We train the Transformer on these (padded) sequences with the log-loss.
Under the (unrealistic) assumptions that the Transformer has the capacity to represent $\hat M_{\cdots}$,
and the learning algorithm can find the representation, 
this (tautologically) implies that the Transformer distribution converges to $\hat M_{\cdots}$.
Similarly if the Transformer is trained on $x^j$ sampled from $M_{s(j),L(j),n}(x)$ for $j=1,2,3,...$,
it converges to $M_{\infty,\infty,n}$. 
For a Transformer with context length $n$ increasing over time,
even $\hat M_{\cdots}\to M$ could be possible.
% limit-normalized M
To guarantee normalized probabilities when learning $\widetilde M_{\cdots}$ and $M_{\cdots}^{norm}$,  
we do \emph{not} introduce a $\bot$-padding symbol.
For $\widetilde M_{\cdots}$ we train on $\widetilde D$ which doesn't require padding.
% Solomonoff normalized M
For training towards $M_{\cdots}^{norm}$, 
we pad the $x^j$ in $D^J$ to length $n$ with arbitrary symbols from $\cX$ and train on that,
but we (have to) cut the log-loss short 
$-\sum_{t=1}^{\ell(x)}\log(\text{LLM}(x_t|x_{<t}))$, i.e.\ $\ell(x)$ rather than $n$,
so as to make the loss hence gradient hence minimum independent of the arbitrary padding.

\paragraph{Limit-normalized $\widetilde M$.}
This is the easiest case: $\widetilde D$ removes strings shorter than $n$ from $D^J$ (sampled from $M$),
so $\widetilde D$ has distribution $\widetilde M$, 
hence for $D=\widetilde D$, the log-loss is minimized by $p_\theta=\hat{\widetilde M}$,
i.e.\ training on $\widetilde D$ makes $p_\theta$ converge to $\hat{\widetilde M}$ (under the stated assumptions).

\paragraph{Unnormalized $M$.}
For this case we need to augment the (token) alphabet $\cX$ with some (absorbing) padding symbol $\bot$:
Let $D_\bot$ be all $x\in D^J$ but padded with some $\bot$ to length $n$.
We can extend $M:\cX^*\to[0;1]$ to $M_{\!\bot\!}:\cX^*\cup\{\bot\}\to[0;1]$ by
\iffalse % compressed
$M_{\!\bot\!}(x):=M(x)$ for $x\in\cX^*$; 
$M_{\!\bot\!}(x\bot^{\!t}):=M(x)-\sum_{a\in\cX}M(xa)$ for $x\in\cX^*$ and $\forall t\geq 1$;
and $M_{\!\bot\!}(x):=0$ for all $x\not\in\cX^*\{\bot\}^*$.
\else % displayed
\begin{align*}
  M_{\!\bot\!}(x)           ~&:=~ M(x)                     & & \text{for all} & & x\in\cX^* \\
  M_{\!\bot\!}(x\bot^{\!t}) ~&:=~ M(x)-\textstyle\sum_{a\in\cX}M(xa) & & \text{for all} & & x\in\cX^* ~~\text{and}~~ t\geq 1 \\
  M_{\!\bot\!}(x)           ~&:=~ 0                        & & \text{for all} & & x\not\in\cX^*\{\bot\}^*
\end{align*}
\fi
It is easy to see that $D_\bot$ has distribution $M_{\!\bot}$,
hence for $D=D_\bot$, the log-loss is minimized by $p_\theta=\hat M_{\!\bot}$.
Since $\hat M_{\!\bot\!}(x)$ restricted to $x\in\cX^*$ is just $\hat M(x)$, 
training on $D_\bot$ makes $p_\theta(x)$ converge to $\hat M(x)$ for $x\in\cX^*$. Though it is possible to train neural models that would converge in the limit to the standard (computable) Solomonoff prior, we focus on the normalized version due to Remark~\ref{rem:inconveniences_semimeasure}. 
\\
\emph{Training variation:}
Note that for $M$, the Transformer is trained to predict $x\bot$ if $\ell(x)<n$.
If $\ell(x)<n$ is due to the time limit $s$ in $U^s$,
it is preferable to \emph{not} train the Transformer to predict $\bot$ after $x$,
since for $s\to\infty$, which we are ultimately interested in,
$x$ may be extended with proper symbols from $\cX$.
One way to achieve this is to cut the log-loss (only) in this case at $t=\ell(x)$ similar to $M^{norm}$ below
to not reward the Transformer for predicting $\bot$.

\subsection{Normalized Solomonoff Loss}\label{sec:normalized_solomonoff_loss}

Here is the derivation of the loss.
\begin{align*}
  \text{Loss}(\theta) ~&:=~ -\frac1J\sum_{x\in D^J}\log p_\theta(x)
  ~=~ -\frac1J\sum_{x\in D^J}\sum_{t=1}^{\ell(x)}\log p_\theta(x_t|x_{<t}) \\
  ~&=~ -\frac1J\sum_{t=1}^n\sum_{x\in D^J\wedge\ell(x)\geq t\hspace{-7ex}}\log p_\theta(x_t|x_{<t})
  ~=~ -\sum_{t=1}^n\sum_{x_{1:t}}\hat M_{D^J}(x_{1:t})\log p_\theta(x_t|x_{<t}) \\
  ~&=~ -\sum_{t=1}^n\sum_{x_{<t}}\Big(\sum_{x_t}\hat M_{D^J}(x_{1:t})\Big)\Big(\sum_{x_t}\hat M^{norm}(x_t|x_{<t})\log p_\theta(x_t|x_{<t})\Big)
\end{align*}
where the last equality follows from (\ref{eq:solnconv}).

%%%%%%%%%%%%%%%%%%%%%%%%%%%%%%%%%%%%%%%%%%%%%%%%%%%%%%%%%%%%%%%
\section{Generalized Solomonoff Semimeasure}\label{sec:generalized_solomonoff_proof}
%%%%%%%%%%%%%%%%%%%%%%%%%%%%%%%%%%%%%%%%%%%%%%%%%%%%%%%%%%%%%%%

\paragraph{Streaming functions.} % 2309(3b)
A streaming function $\varphi$ takes a growing input sequence and produces a growing output sequence.
In general, input and output may grow unboundedly or stay finite.
Formally, $\varphi:\cX^\#\to\cX^\#$, where $\cX^\#:=\cX^\infty\cup\cX^*$.
In principle input and output alphabet could be different,
but for simplicity we assume that all sequences are binary, i.e.\ $\cX=\{0,1\}$.
For $\varphi$ to qualify as a streaming function,
we need to ensure that extending the input only extends and does not modify the output.
Formally, we say that 
\begin{align*}
  \varphi ~\text{is monotone ~~~ iff}~~~ [\forall q\sqsubseteq p: \varphi(q)\sqsubseteq\varphi(p)]
\end{align*}
where $q\sqsubseteq p$ means that $q$ is a prefix of $p$ i.e. $\exists r\in\cX^\#:qr=p$, and $\sqsubset$ denotes strict prefix $r\neq\epsilon$.
$p$ is $\varphi$-minimal for $x$ if $\exists r:\phi(p)=xr$ and $\forall r\forall q\sqsubset p:\phi(q)\neq xr$.
We will denote this by $\varphi(p)=x*$. $p$ is the shortest program outputting a string starting with $x$.

\paragraph{Monotone Turing Machines (MTM).} % 2309(3b)
A Monotone Turing machine $T$ is a Turing machine with left-to-right read-only input tape,
left-to-right write-only output tape, and some bidirectional work tape.
The function $\varphi_T$ it computes is defined as follows:
At any point in time after writing the output symbol but before moving the output head 
and after moving the input head but before reading the new cell content,
if $p$ is the content left of the current input tape head,
and $x$ is the content of the output tape up to the current output tape head,
then $\varphi_T(p):=x$. It is easy to see that $\varphi_T$ is monotone.
We abbreviate $T(p)=\varphi_T(p)$.
There exist (so called optimal) universal MTM $U$ that can emulate any other MTM via
$U(i'q)=T_i(q)$, where $T_1,T_2,...$ is an effective enumeration of all MTMs 
and $i'$ a prefix encoding of $i$ \citep{hutter2004universal,li2019introduction}.

\subsection{Proof of Theorem~\ref{thm:ugsol}}

\ugsol*

% See 2309(3f) for proof
We can effectively sample from any computable $Q$ if we have access to infinitely many fair coin flips.
The conditions on $Q$ ensure that the entropy of $Q$ is infinite, 
and stays infinite even when conditioned on any $q\in\cX^*$.
This also allows the reverse: Converting a sample from $Q$ into infinitely many uniform random bits.
Forward and backward conversion can be achieved sample-efficiently via (bijective) arithmetic (de)coding.
This forms the basis of the proof below.
% See 2309(3c) for proof
The condition of $Q$ being a proper measure rather than just being a semimeasure is also necessary:
For instance, for $Q(q)=4^{-\ell(q)}$, on a Bernoulli($\frac12$) sequence $x_{1:\infty}$, $M_U(x_t|x_{<t})\to\frac12$ as it should,
one can show that $M_U^Q(x_t|x_{<t})<\frac13$ for infinitely many $t$ (w.p.1).

\begin{proof} \emph{(sketch)} % 2309(3e)
Let $0.q_{1:\infty}\in[0;1]$ be the real number with binary expansion $q_{1:\infty}$. 
With this identification, $Q$ can be regarded as a probability measure over $[0;1]$.
Let $F:[0;1]\to[0;1]$ be its cumulative distribution function,
which can explicitly be represented as $F(0.q_{1:\infty})=\sum_{t:q_t=1}Q(\Gamma_{q_{<t}0})$,
since $[0;\,0.q_{1:\infty})={\dot{\smash{\bigcup}}}_{t:q_t=1}0.\Gamma_{q_{<t}0}$, 
where $0.\Gamma_q=[0.q0^\infty;\,0.q1^\infty)$ and $\dot{\smash{\bigcup}}$ denotes disjoint union.
Now assumption $Q(q)>0~\forall q\in\cX^*$ implies that $F$ is strictly increasing,
and assumption $Q(q_{1:n})\to 0$ implies that $F$ is continuous.
Since $F(0)=0$ and $F(1)=1$, this implies that $F$ is a bijection.
Let $0.p_{1:\infty}=F(0.q_{1:\infty})$ and $0.q_{1:\infty}=F^{-1}(0.p_{1:\infty})$.
\footnote{Note that $p_{1:m}$ is uniformly distributed and is (for some $m$) essentially the arithmetic encoding of $q_{1:n}$ with one caveat: 
The mapping from sequences to reals conflates $0.q10^\infty=0.q01^\infty$.
Since the set of all conflated sequences has probability $0$,
(under $Q$ as well as Bernoulli($\frac12$)), any error introduced due to this conflation
has no effect on the distribution $M_U^Q(x)$.}.
Further for some finite prefix $q\sqsubset q_{1:\infty}$, 
we partition the interval 
\begin{align*}
  [0.p_{1:\infty}^0;\,0.p_{1:\infty}^1) ~:=~ [F(0.q0^\infty);F(0.q1^\infty)) ~=:~ \dot\bigcup_{p\in\Phi(q)} 0.\Gamma_p
\end{align*}
into a minimal set of binary intervals $0.\Gamma_p$, 
where $\Phi(q)$ is a minimal prefix free set in the sense that for any $p$, at most one of $p$, $p0$, $p1$ is in $\Phi(q)$.
An explicit representation is
\begin{align*}
  \Phi(q) ~:=~ \{p_{<t}^0 1:t>t_0\wedge p_t^0=0\}~\dot\cup~\{p_{<t}^1 0:t>t_0\wedge p_t^1=1\}
\end{align*}
where $t_0$ is the first $t$ for which $p_t^0\neq p_t^1$. Now we plug
\begin{align*}
  Q(q) ~&=~ F(0.q1^\infty)-F(0.q0^\infty) ~=~ \sum_{p\in\Phi(q)}|0.\Gamma_p| ~=~ \sum_{p\in\Phi(q)}2^{-\ell(p)} ~~~\text{into} \\
  M_U^Q(x) ~&\equiv~ \sum_{q:U(q)=x*\!\!\!\!}Q(q) ~=~ \sum_{\!\!\!\!q:U(q)=x*~~}\sum_{p\in\Phi(q)\!\!\!\!}2^{-\ell(p)} ~=~ \sum_{p:V(p)=x*\!\!\!\!}2^{-\ell(p)} ~=~ M_V(x)
\end{align*}
where $V(p):=U(q)$ for the maximal $q$ such that $p\in\Phi(q)$. 
The maximal $q$ is unique, since $\Phi(q)\cap\Phi(q')=\{\}$ if $q\not\sqsubseteq q'$ and $q'\not\sqsubseteq q$, and finite since $F$ is continuous.

% V is universal
It remains to show that $V$ is universal. 
Let $p^\imath$ be such that $0.\Gamma_{p^\imath}\subseteq[F(0.\imath'0^\infty);F(0.\imath'1^\infty))$.
The choice doesn't matter as long as it is a computable function of $\imath$, but shorter is ``better''.
This choice ensures that $F^{-1}(0.p^\imath*)=0.\imath'...$ whatever the continuation $*$ is.
Now let $F(q_{1:\infty})_\text{tail}:=F(q_{1:\infty})_{\ell(p^\imath)+1:\infty}=p_{\ell(p^\imath)+1:\infty}$ if $q_{1:\infty}$ starts with $\imath'$,
and arbitrary, e.g.\ $F(q_{1:\infty})$, otherwise.
Let $T$ be a MTM with $T(q_{1:\infty}):=U_0(F(q_{1:\infty})_\text{tail})$ for some universal MTM $U_0$.
By Kleene's 2nd recursion theorem \citep[Chp.6]{Sipser:12}, 
there exists an $i$ such that $T_i(q)=T(i' q)~\forall q$.
Let $\dot k:=\ell(i')+1$ and $\dot\ell:=\ell(p^i)+1$ and $q_{<\dot k}:=i'$, hence $p_{<\dot\ell}=p^i$.
Now $V(p_{1:\infty})=U(q_{1:\infty})$ implies
\begin{align*}
  V(p^i p_{\dot\ell:\infty}) ~=~ U(i' q_{\dot k:\infty}) ~=~ T_i(q_{\dot k:\infty}) ~=~ T(i' q_{\dot k:\infty}) ~=~ U_0(F(i' q_{\dot k:\infty})_\text{tail}) ~=~ U_0(p_{\dot\ell:\infty})
\end{align*}
hence $V$ is universal, which concludes the proof.
\end{proof}

\paragraph{Practical universal streaming functions.}
Turing machines are impractical and writing a program for a universal streaming function is another layer of indirection which is best to avoid.
Programming languages are already universal machines.
We can define a conversion of real programs from/to binary strings and prepend it to the input stream.
When sampling input streams $q_{1:\infty}$ we convert the beginning into a program of the desired programming language,
and feed it the tail as input stream.

%-------------------------------------------------------------%

%%%%%%%%%%%%%%%%%%%%%%%%%%%%%%%%%%%%%%%%%%%%%%%%%%%%%%%%%%%%%%%
\section{Experiment methodology details}
%%%%%%%%%%%%%%%%%%%%%%%%%%%%%%%%%%%%%%%%%%%%%%%%%%%%%%%%%%%%%%%
\subsection{Architecture details}\label{sec:architecture_details}

\begin{table}[ht!]
  \caption{Architectures }
  \label{tab:architecture}
  \begin{center}
    \begin{tabular}{llll}
    \toprule
    \textbf{RNN and LSTMs} & \textbf{S} & \textbf{M} & \textbf{L} \\
    \midrule
     RNN Hidden size            & 16 & 32 & 128 \\
     Number of RNN layers   & 1 & 2 & 3 \\
     MLP before RNN layers  & (16,) & (32, 32) & (128, 128, 128)\\
     MLP after RNN layers   & (16,) & (32, 32) & (128, 128, 128)\\
    \midrule
    \textbf{Transformer SINCOS} & & & \\
     Embedding dimension        & 16 & 64 & 256 \\
     Number of heads            & 2 & 4 & 4 \\
     Number of layers            & 2 & 4 & 6 \\
    \bottomrule
\end{tabular}
  \end{center}
\end{table}

\paragraph{RNN.} A vanilla multi-layer RNN~\citep{elman1990finding} with hidden sizes and multi-layer perceptron (MLP) before and after the RNN layers as described in Table~\ref{tab:architecture}.

\paragraph{Stack-RNN.}
A multi-layer RNN controller with hidden sizes and MLP exactly the same as the RNN and LSTMs on Table~\ref{tab:architecture} with access to a differentiable stack~\citep{joulin2015inferring}.
The controller can perform any linear combination of \push{}, \pop{}, and \noop{} on the stack of size according to Table~\ref{tab:architecture}, with action weights given by a softmax over a linear readout of the RNN output. Each cell of the stack contains a real vector of dimension 6 and the stack size is 64 for all (S, M and L) sizes.

\paragraph{Tape-RNN.}
A multi-layer RNN controller with hidden sizes according to the Table~\ref{tab:architecture} with access to a differentiable tape, inspired by the Baby-NTM architecture~\citep{suzgun2019memory}.
The controller can perform any linear combination of \writeright{}, \writeleft{}, \writestay{}, \jumpleft{}, and \jumpright{} on the tape, with action weights given by a softmax.
The actions correspond to: writing at the current position and moving to the right (\writeright{}), writing at the current position and moving to the left (\writeleft{}), writing at the current position (\writestay{}), jumping $\ell$ steps to the right without writing (\jumpright{}), where $\ell$ is the length of the input, and jumping $\ell$ steps to the left without writing (\jumpleft{}).
% see \cref{sec:experimental-details:problem-setup} below)
As in the Stack-RNN, each cell of the tape contains a real vector of dimension 6 and the tape size is 64 for all (S, M and L) sizes.

\paragraph{LSTM.}
A multi-layer LSTM~\citep{hochreiter1997long} of hidden sizes according to Table~\ref{tab:architecture}.

\paragraph{Transformer decoder.}
A vanilla Transformer decoder~\citep{vaswani2017attention}. See Table~\ref{tab:architecture} for the embedding dimension, number of heads and number of layers for each model size (S, M and L). Each layer is composed of an attention layer, two dense layers, and a layer normalization. We add a residual connections as in the original architecture~\citep{vaswani2017attention}.
We consider the standard sin/cos~\citep{vaswani2017attention} positional encoding.

%-------------------------------------------------------------%
\subsection{CTW} \label{sec:ctw_details}
%-------------------------------------------------------------%
\def\frs#1#2{{^{#1}\!/\!_{#2}}} % ^#1/_#2 fraction

Below is an ultra-compact introduction to (sampling from) CTW \citep{willems1995context,willems1997reflections}.
For more explanations, details, discussion, and derivations, see \citep[Chp.4]{Hutter:24uaibook2}.

\paragraph{A variable-order Markov process.}
is a probability distribution over (binary) sequences $x_1,x_2,x_3,...$ with the following property:
Let $S\subset\{0,1\}^*$ be a complete suffix-free set of strings (a reversed prefix-free code) which can equivalently be viewed as a perfect binary tree.
Then $p(x_t=0|x_{<t};S,\Theta_S):=\theta_s$ if (the unique) context of $x_t$ is $s=x_{t-\ell(s):t-1}\in S$, and $\Theta_S:=(\theta_s\in[0;1]:s\in S)$.
%Then $p(x_t=0|x_{t-\ell(s):t-1}=s,S,\Theta):=\theta_s$ for (the unique) $s\in S$, where $\theta_s\in[0;1]$.
We arbitrarily define $x_t=0$ for $t\leq 0$.

\paragraph{Intuition about Variable-order Markov sources} VOMS considers data generated from tree structures. For example, given the binary tree 
\begin{verbatim}
                Root
             0/      \1
        Leaf_0        Node
                    0/     \1
              Leaf_10       Leaf_11
\end{verbatim}

and given the history of data “011“ (where 0 is the first observed datum and 1 is the last one) the next sample uses $\text{Leaf}_{11}$ (because the last two data points in history were 11) to draw the next datum using a sample from a Beta distribution with parameter $\text{Leaf}_{11}$. Say we sample a 0, thus history is then transformed into “0110” and $\text{Leaf}_{10}$ will be used to sample the next datum (because now the last two datapoints that conform to a leaf are "10"), and so forth.  This way of generating data is very general and can produce many interesting patterns ranging from simple regular patterns like  $01010101$ or more complex ones that can have stochastic samples in it. Larger trees can encode very complex patterns indeed. 

\paragraph{Sampling from CTW.}
Context Tree Weighting (CTW) is a Bayesian mixture over all variable-order Markov sources of maximal order $D\in\mathbb N_0$,
i.e.\ over all trees $S$ of maximal depth $D$ and all $\theta_s\in[0;1]$ for all $s\in S$.
The CTW distribution is obtained as follows:
\iftrue % short version
We start with an empty (unfrozen) $S=\{\epsilon\}$. 
Recursively, for each unfrozen $s\in S$ with $\ell(s)<D$, 
with probability $\frs12$ we freeze $s$; 
with probability $\frs12$ we split $S\leftarrow S\setminus\{s\}\cup\{0s,1s\}$ 
until all $s\in S$ are frozen or $\ell(s)=D$.
\else % long version
Let $S$ be a tentative suffix set, some $s\in S$ marked with $\sim$ as tentative.
We start with $S=\{\tilde\epsilon\}$, where $\epsilon$ is the empty string.
Recursively, for each tentative $\tilde s\in S$ with $\ell(s)<D$, 
with probability $\frs12$, we freeze it $S\leftarrow S\setminus\{\tilde s\}\cup\{s\}$; 
with probability $\frs12$ we split it $S\leftarrow S\setminus\{\tilde s\}\cup\{\widetilde{0s},\widetilde{1s}\}$ 
until all $s\in S$ are frozen or $\ell(s)=D$ (and remove the $\sim$).
\fi
Then we sample $\theta_s$ from $\text{Beta}(\frs12,\frs12)$ for all $s\in S$.
Finally for $t=1,2,3,...$ we sample $x_t$ from $p(x_t|x_{<t};S,\Theta_S)$.

\paragraph{Computing CTW.}
The CTW probability $P_\text{CTW}(x_{1:n})$ can be calculated as follows:
Let $a_s:=|\{t\in\{1,...,n\}:x_t=0\wedge x_{t-\ell(s):t-1}=s\}|$ count the number of $x_t=0$ immediately preceded by context $s\in\{0,1\}^*$,
and similarly $b_s:=|\{t:x_t=1\wedge x_{t-\ell(s):t-1}=s\}|$. 
Let $x_{1:n}^s\in\{0,1\}^{a_s+b_s}$ be the subsequence of $x_t$'s that have context $s$.
For given $\theta_s$ for $s\in S$, $x_{1:n}^s$ is i.i.d.\ (Bernoulli($1-\theta_s$)).
Hence for $\theta_s\sim \text{Beta}(\frs12,\frs12)$,
$P(x_{1:n}^s|s\in S)=P_\text{KT}(a_s,b_s):=\int_0^1 \theta_s^{a_s}(1-\theta_s)^{b_s}\text{Beta}(\frs12,\frs12)(\theta_s)d\theta_s$.
If $s\not\in S$, we split $x_{1:n}^s$ into $x_{1:n}^{0s}$ and $x_{1:n}^{1s}$.
By construction of $S$, a tentative $s\in S$ gets replaced by $0s$ and $1s$ with 50\% probability, recursively,
hence 
$P_\text{CTW}(x_{1:n}^s)=\frac12 P_\text{KT}(a_s,b_s) + \frac12 P_\text{CTW}(x_{1:n}^{0s})P_\text{CTW}(x_{1:n}^{1s})$ terminating with 
$P_\text{CTW}(x_{1:n}^s)=P_\text{KT}(a_s,b_s)$ when $\ell(s)=D$.
This completes the definition of $P_\text{CTW}(x_{1:n})\equiv P_\text{CTW}(x_{1:n}^\epsilon)$.
Efficient $O(nD)$ algorithms for computing $P_\text{CTW}(x_{1:n})$ 
(and updating $n\to n+1$ in time $O(D)$) 
and non-recursive definitions
can be found in \citet[Chp.4]{Hutter:24uaibook2}.

\paragraph{Distributions of Trees.} % 2309(2)
A tree has depth $\leq d$ if either it is the empty tree 
or if both its subtrees have depth $<d$.
Therefore the probability of sampling a tree of depth $\leq d$ is
$F(d)=\frac12+\frac12 F(d-1)^2$, with $F(0)=\frac12$. 
Therefore the probability of sampling a tree of depth $d$ is
$P(d)=F(d)-F(d-1)$ for $d<D$ and $P(D)=1-F(D-1)$.
The theoretical curve ($P(0)=\frac12$, $P(1)=\frac18$, $P(2)=\frac9{128}$,...) 
is plotted in Fig.~\ref{fig:num_datapoints_vs_tree_depth}  together with the empirical distribution.
More meaningful is probably the expected number of leaf nodes at each level $d$.
Since each node at level $d$ is replaced with prob.\ $\frac12$ by two nodes at level $d+1$,
the expected number of leaf nodes $E(d)$ is the same at all levels $d<D$.
Since $E(0)=\frac12$, we have $E(d)=\frac12$ for all $d<D$ and $E(D)=1$,
hence the total expected number of leaf nodes is $E_+=\frac12 D+1$.
While this doesn't sound much, it ensures that for $N=10'000$ samples,
we uniformly test $5'000$ contexts for each length $d<D$.
We can get some control over the distribution of trees by splitting nodes at level $d$ with probability $\alpha_d\in[0;1]$ instead of $\frac12$.
In this case, 
$E(d)=2\alpha_0\cdot...\cdot 2\alpha_{d-1}(1-\alpha_d)$ for $d<D$.
%$E(d)=(1-\alpha_d)\prod_{a=0}^{d-1}(2\alpha_a)$ for $d<D$.
So for $\alpha_d>\frac12$ we can create trees of size exponential in $D$,
and (within limits) any desired depth distribution.

%-------------------------------------------------------------%
\subsection{Chomsky}\label{sec:chomsky_method_details}
%-------------------------------------------------------------%

\begin{table}[ht!]
  \caption{
    Table taken from \citep{deletang2022neural}. Tasks with their level in the Chomsky hierarchy and example input/output pairs.
    The $\dagger$ denotes permutation-invariant tasks; the $\star$ denotes counting tasks; the $\circ$ denotes tasks that require a nondeterministic controller; and the $\times$ denotes tasks that require superlinear running time in terms of the input length.
  }
  \label{tab:tasks}
  \begin{center}
    \resizebox{\columnwidth}{!}{
    \begin{tabular}{@{}llll@{}}
    \toprule
    \textbf{Level} & \textbf{Name} & \textbf{Example Input} & \textbf{Example Output} \\
    \midrule
    \multirow{4}{*}{Regular (R)} 
    & \evenpairs & $aabba$ & True \\
    & \modulararithmeticsimple & $1+2-4$ & $4$ \\
    & \paritycheck$^\dagger$ & $aaabba$ & True \\
    & \cyclenavigation$^\dagger$ & $011210$ & $2$ \\
    \\
    % & \compareoccurrence & $aabba$ & True \\
    \multirow{4}{*}{Deterministic context-free (DCF)}
    & \stackmanipulation & $abbaa$ \pop{} \push{} $a$ \pop{} & $abba$ \\
    & \reversestring & $aabba$ & $abbaa$ \\
    & \modulararithmeticbrackets & $-(1-2)\cdot(4-3\cdot(-2))$ & $0$ \\
    & \solveequation$^\circ$ & $-(x-2)\cdot(4-3\cdot(-2))$ & $1$ \\
    \\
    \multirow{7}{*}{Context-sensitive (CS)}
    & \duplicatestring & $abaab$ & $abaababaab$ \\
    & \missingduplicate & $10011021$ & $0$ \\
    & \oddsfirst & $aaabaa$ & $aaaaba$ \\
    & \binaryaddition & $10010+101$ & $10111$ \\
    & \binarymultiplication$^\times$ & $10010*101$ & $1001000$ \\
    & \computesqrt & $100010$ & $110$ \\
    & \bucketsort$^{\dagger\bigstar}$ & $421302214$ & $011222344$ \\
    \bottomrule
\end{tabular}
}
  \end{center}
\end{table}

%-------------------------------------------------------------%
\section{UTMs: Brainf*ck and BrainPhoque} \label{sec:utm_methods_details}
%-------------------------------------------------------------%

Our BrainPhoque (BP) UTM produces program evaluation traces that are equivalent to those of brainf*ck (BF) programs~\citep{brainfck} (see also $\mathcal{P}''$~\citep{bohm1966pprimeprime}), but the programs are written slightly differently:
they are even less human-readable but have better properties when sampling programs.

We start by giving a quick overview of the BF machine, then explain why we need a slightly different machine, and its construction. Finally we explain how to shorten sampled programs and calculate an upper bound on the log-loss.

\begin{figure}
    \centering
    \includegraphics[width=\textwidth]{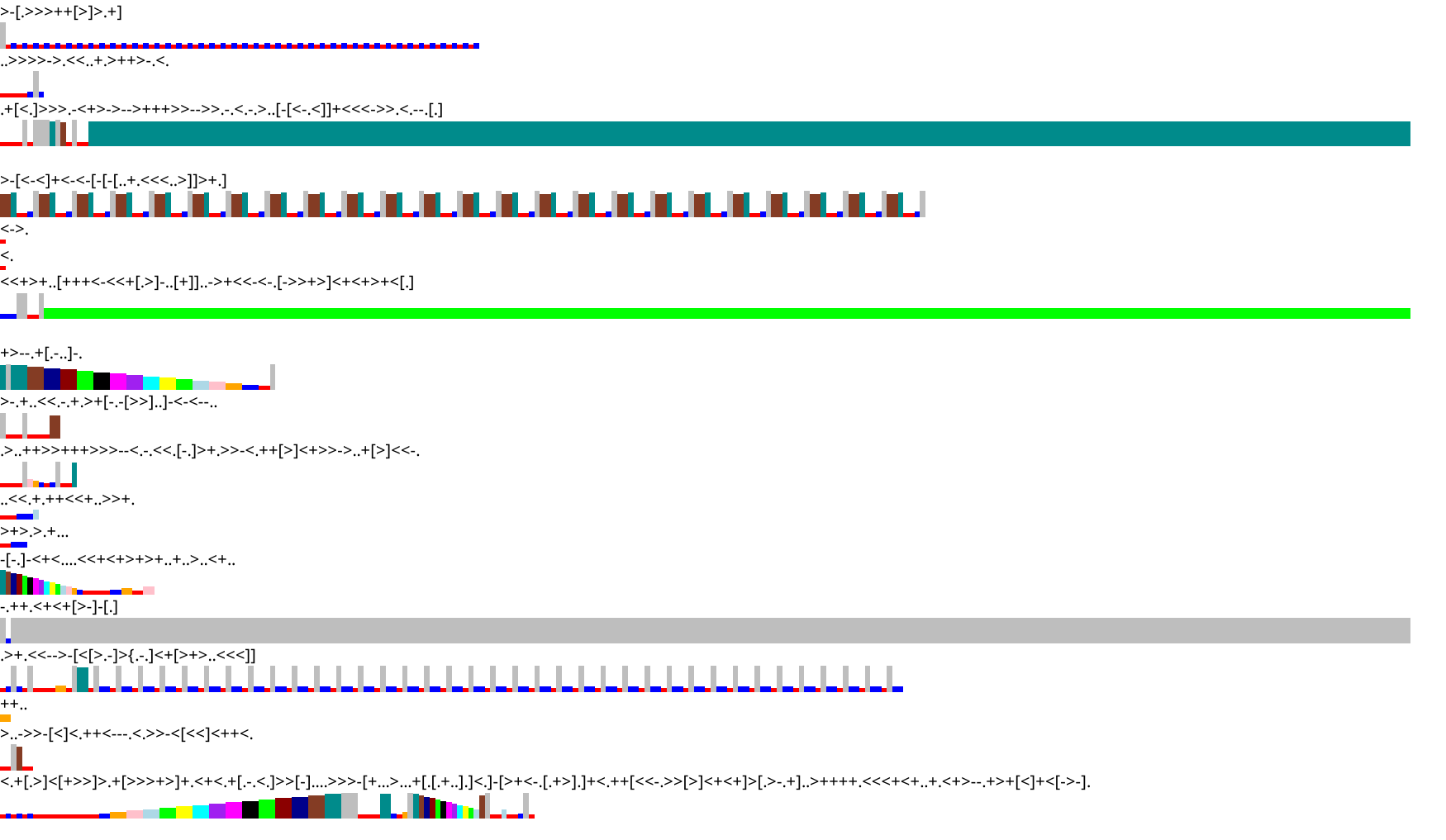}
    \caption{Some BrainPhoque programs and their corresponding outputs (truncated at 256 symbols). The smallest bars (in red) correspond to the value 0, and the largest bars (in gray) correspond to value 16.
    The programs have been reduced after evaluation by removing 
    a set of unnecessary instructions.
    Most of the generated outputs are regular, and only about 1 in $5000$ sampled programs exhibits non-regular patterns.
    But see Table \ref{tab:mcsampling} for a way to improve these numbers and generate more interesting and complex sequences.}
    \label{fig:brainphoque}
\end{figure}

See Figure \ref{fig:brainphoque} for some sample programs and outputs.

\subsection{Brainf*ck}

BF is one of the smallest and simplest Turing-complete programming languages.
It features a read-only input tape, a working tape, and a write-only output tape. These tapes are assumed infinite
but for practical purposes they are usually fixed at a finite and constant length and initialized with 0.\footnote{The tape could also grow on request, but this tends to slow down program evaluation.}
Each tape cell can contain a non-negative integer, which can grow as large as the 'alphabet size'. Above that number, it loops back to 0. In the paper, we choose an alphabet size of 17.

Each tape has a pointer. For simplicity, the pointer of the working tape is called WTP, and the value at the WTP is called \emph{datum}, which is an integer.

BF uses 8 instructions \bfcode{<>+-[],.} which are:
\begin{itemize}
    \item \bfcode{<} and \bfcode{>} decrement and increment the WTP, modulo the length of the tape.
    \item \bfcode{+} and \bfcode{-} increment and decrement the datum,
    modulo the alphabet size.
    \item \bfcode{[} is a conditional jump: if the datum is 0,
    the instruction pointer jumps to the corresponding (matching) \bfcode{]}.
    \item \bfcode{]} is an unconditional jump to the corresponding \bfcode{[}.%
    \footnote{For efficiency reasons the instruction \bfcode{]} is usually defined to jump to the matching \bfcode{[} if the datum is non-zero. We stick to a unconditional jump for simplicity reasons.}
    \item \bfcode{,} copies the number under the reading tape pointer into the datum cell, and
    increments the reading pointer.
    \item \bfcode{.} copies the datum to the output tape at the output pointer and increments the output pointer.
\end{itemize}

In this paper we do not use an input tape, so we do not use the \bfcode{,} instruction.

When evaluating a program, the instruction pointer is initially
on the first instruction, the output tape is empty, and the working tape is filled with zeros.
Then the instruction under the instruction pointer is evaluated according to the above rules, and the instruction pointer is moved to the right.
Evaluation terminates when the number of evaluated instructions reaches a given limit,
or when the number of output symbols reaches a given limit.

For a sequence of instructions \bfcode{A[B]C}, where \bfcode{A}, \bfcode{B} and \bfcode{C} are sequences of (well-balanced) instructions, we call \bfcode{B} the \emph{body} of the block
and \bfcode{C} the \emph{continuation} of the block.

\subsection{BrainPhoque: Simultaneous generation and evaluation}

We want to sample arbitrary BF programs and evaluate them for $T$ steps each.
To maximize computation efficiency of the sampling and running process, programs containing unbalanced parentheses are made valid, in particular by skipping any additional \bfcode{]}.
\todo{brackets are matched moste-nested first, that is, in\bfcode{[[]} the closing bracket is matched with the second opening bracket.}

Since we want to approximate \emph{normalized} Solomonoff induction \ref{def:normalized_solomonoff_prior}, we can make a few simplifications.
In particular, programs do not need to halt explicitly, which removes the need for a halting symbol and behaviour.%
\footnote{The halting behaviour can be recovered by ending programs with a particular infinite loop such as \bfcode{[]+[]} (which loops whether the datum is zero or not), and terminate the evaluation (instead of looping forever) upon evaluating this sequence.}
Hence we consider that \emph{all} programs are infinite, but that at most $T$ instructions are evaluated.
The difficulty with BF programs is that the evaluated instructions can be at arbitrary locations on the program tape, since large blocks \bfcode{[...]} may be entirely skipped, complicating both the sampling process and \XXX

This can be fixed by generating BF programs as trees, where  branching on opening brackets \bfcode{[}:
The left branch corresponds to the body of the block (and terminates with a \bfcode{]}),
while the right branch corresponds to the continuation of the block.
When encountering an opening bracket for the first time during evaluation,
which branch is evaluated next depends on the datum.
Hence, to avoid generating both branches, we need to generate the program \emph{as it is being evaluated}:
when sampling and evaluating a \bfcode{[}, if the datum is 0 we follow the right branch and start sampling the continuation without having to sample the body (for now); conversely, if the datum is not zero, we follow the left branch and start sampling and evaluating the continuation.
If the same opening bracket is later evaluated again with a different datum value, the other branch may be generated and evaluated.

Our implementation of program generation and evaluation in BrainPhoque uses one growing array for the program, one jump table, and one stack for yet-unmatched open brackets.

If the instruction pointer is at the end of the program,
a new instruction among \bfcode{+-<>[].} is sampled;
if it is \bfcode{[} and the datum is 0, it is changed to \bfcode{\{}.
The new instruction is appended to the program,
and is then evaluated.
If the new instruction is \bfcode{[}, the next instruction to be sample (and appended to the program) is the beginning of the body of the block,
but if instead the new instruction is \bfcode{\{}, the next instruction to be sampled (and appended to the program) is the continuation of the body.
At this point the jump table does not yet need to be updated --- since the next instruction to evaluate is also the next instruction in location.
The jump table is updated to keep track of where the continuations and bodies are located in the program.
If the instruction pointer eventually comes back for a second time of an opening bracket \bfcode{[} (resp. \bfcode{\{}) and the datum is now 0 (resp. not 0), the continuation (resp. body) of the block must now be sampled and appended to the program; and now the jump table must be updated accordingly.

The stack of unmatched brackets is updated only when the body of a block is being generated.

Some properties of BrainPhoque:
\begin{itemize}
    \item If a program is run for $t+k$ steps, it behaves the same on the first $t$ steps for all values of $k$.\footnote{While this is an obviously desirable property, it is also easy to overlook.}
    In particular, unmatched opening brackets behave the same whether they will be matched or not.
    \item Program generation (sampling) only requires a single growing-only array. A tree structure is not required. This is the reason for having the additional \bfcode{\{} instruction, which makes it clear --- once evaluated the second time --- whether the body or the continuation has already been generated.
    % \item Even though BrainPhoque programs uses an additional instruction compared to brainf*ck, when generating a program the instructions \bfcode{[} and \bfcode{\{} are mutually exclusive (depending on the datum) and thus only 7 instructions need to be sampled.
    \item If the instruction pointer is at cell $n$, then all instructions to the left of $n$ have been evaluated at least once.
    If this is the first evaluation of cell $n$, then no instruction to the right of $n$ have been evaluated yet.
\end{itemize}

\iffalse % horizontal table 
\begin{table}
\includegraphics[width=\textwidth]{assets/markov-chain-order2-filtered-programs-wide.pdf}%
\caption{\bf Pre-trained BP program sampling probabilities}
\end{table}
\fi

\begin{table}
\begin{minipage}[c]{0.49\textwidth}
%\resizebox{0.95\columnwidth}{!}{\input{context-table}}
\includegraphics[height=0.9\textheight]{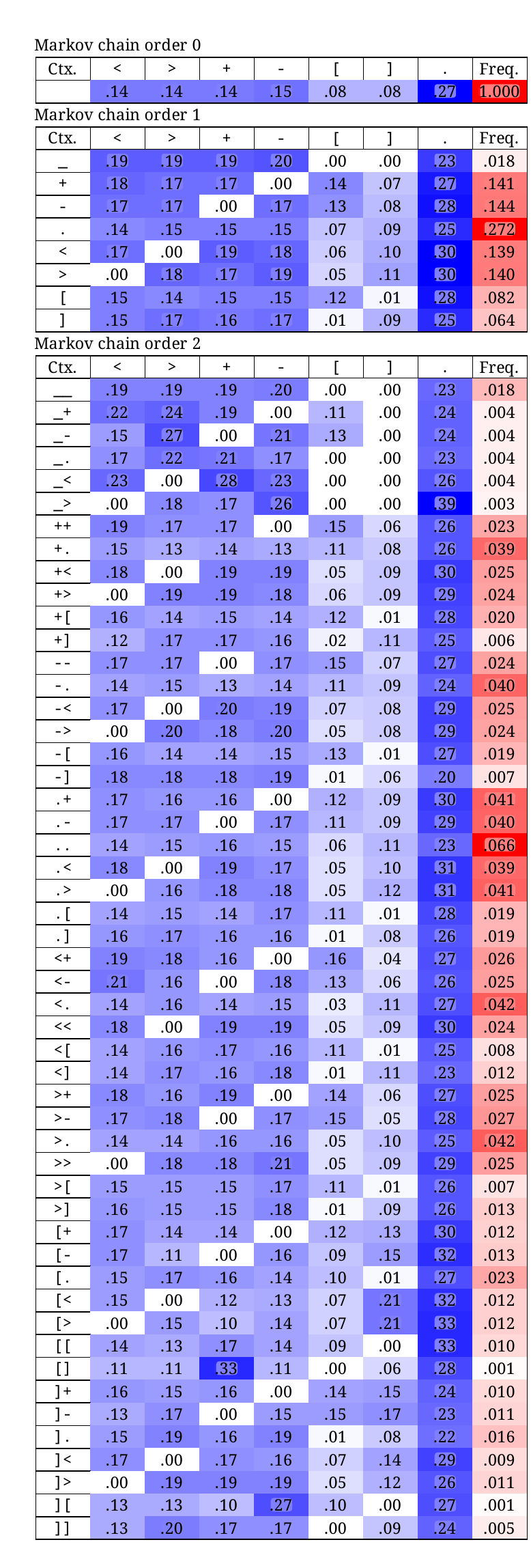}
\end{minipage}
\vspace{-3ex}
\hspace{1ex}
\begin{minipage}[c]{0.49\textwidth}
\caption{
{\bf Pre-trained BP program sampling probabilities}
Instead of sampling programs uniformly, we can sample them w.r.t.\ any
probability distribution $Q$ that satisfies Theorem~\ref{thm:ugsol}.
We initially sampled programs uniformly and filtered out `boring' sequences.
Then we trained $Q$ via cross-entropy to mimic the distribution of `interesting' sequences.
We used a 2nd-order Markov process as a model for $Q$.
While uniform sampling resulted in only 0.02\% interesting sequences,
sampling from $Q$ increased it to 2.5\%, a 137-fold improvement.
The table on the left shows the 0th, 1st, and 2nd order Markov processes
$Q(p_t)$,
$Q(p_t|p_{t-1})$, and
$Q(p_t|p_{t-2}p_{t-1})$ from which BP programs are sampled, for $p_\cdot\in\{\bfcode{<>+-[]\{.}\}$,
but where results for \bfcode{[} and \bfcode{\{} have been merged.
Each row corresponds to a context (none or $p_{t-1}$ or $p_{t-2}p_{t-1}$). 
We also included $Q(p_1|p_0\!\!:=\!\!\bfcode{\_})$ and $Q(p_1|p_{-1}p_0\!\!:=\!\!\bfcode{\_\_})$.
The entries in each column correspond to the sampling probability of $p_t$ in the corresponding row-context.
Training on interesting sequences has led to a non-uniform distribution $Q$. Universality is preserved for any $k$-order Markov process, provided all transition probabilities are non-zero.
The probability $Q(\bfcode{.})$ of outputting a symbol has nearly doubled from 0.14 to 0.27 on average,
while the probability of loop brackets ($\bfcode{[,]}$) reduced to 0.07 each on average.
The marginal probabilities $Q(\bfcode{<})\approx Q(\bfcode{>})\approx Q(\bfcode{+})\approx Q(\bfcode{-})\approx 1/7$ have not changed much,
but many of the conditional ones have.
Certain combination of instructions are now blocked: 
For instance \bfcode{+-} and \bfcode{-+} and \bfcode{<>} and \bfcode{><} have probability close to $0$,
since they cancel each other and hence are redundant.
Some triples such as \bfcode{][-} and \bfcode{<+} and \bfcode{>-} and others are enhanced.
\\
Caveat: We did not have time to retrain our NN models on these newly generated sequences (experiments are still running).
But since the statistics is improved, we expect the results in Figures~\ref{fig:main_results_utm} and \ref{fig:utm_to_chomsky_transfer} to improve or at least not deteriorate.
%\vspace*{13ex} 
}
\label{tab:mcsampling}
\end{minipage}
\end{table}

\subsection{Solomonoff log-loss upper bound and shortening programs}

We tried to provide a meaningful upper bound for the loss of Solomonoff induction for Figure \ref{fig:main_results_utm}, but this is far from easy. 
See Section~\ref{sec:methods} for context.
As mentioned there, to calculate a more meaningful upper bound,
we shorten programs by recursively removing unnecessary
open brackets and closing brackets that are unmatched,
as well as all self-cancelling pairs of instructions
(\bfcode{+-}, \bfcode{-+}, \bfcode{<>},\bfcode{><}).
Moreover, we remove all instructions of the program that have been
evaluated for the first time after the last evaluation of a print \bfcode{.} instruction (since they do not participate in producing the output.
This procedure often reduces programs by a third.
Programs that do not output anything are thus reduced to the empty program (probability 1).

If $q$ is a sampled program, then $\tilde q$ is the corresponding shortened program.
We calculate an upper bound on the loss of the Solomonoff predictor,
with U = BrainPhoque,
on a set of sampled programs $\hat Q = (q^1,\dots,q^J)$ and corresponding outputs
$(U(q^1)_{1:256}, \dots, U(q^J)_{1:256})$,
\begin{align}
    \text{Loss}(M_U, \hat Q) = \sum_{q\in \hat Q} -\log \sum_{p: U(p)_{1:256}=U(q)_{1:256}} 7^{-\ell(p)}
    \leq \sum_{q\in \hat Q} -\log  7^{-\ell(\tilde q)}
    = \log(7)\sum_{q\in \hat Q} \ell(\tilde q)
\end{align}
since the program alphabet is not binary but has $7$ instructions.
Unfortunately, even after reduction this bound is still quite loose, but improving this bound meaningfully would likely require a much larger amount of computation.

\section{Additional Results Details}
\begin{figure*}[ht!]
    \centering
    \begin{subfigure}[t]{0.45\textwidth}
        \centering
        \includegraphics[width=0.65\textwidth]{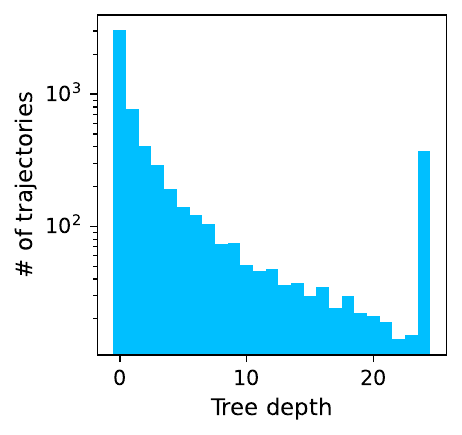}
        \caption{Tree depth per trajectory.}
        \label{fig:num_datapoints_vs_tree_depth}
    \end{subfigure}
    \hfill
    \begin{subfigure}[t]{0.45\textwidth}
        \centering
        \includegraphics[width=0.65\textwidth]{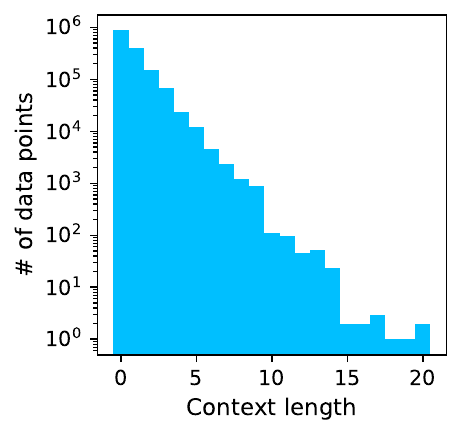}
        \caption{Context length per datapoint.}
        \label{fig:num_datapoints_vs_context_length}
    \end{subfigure}
    \begin{subfigure}[b]{\textwidth}
        \centering
        \includegraphics[width=\textwidth]{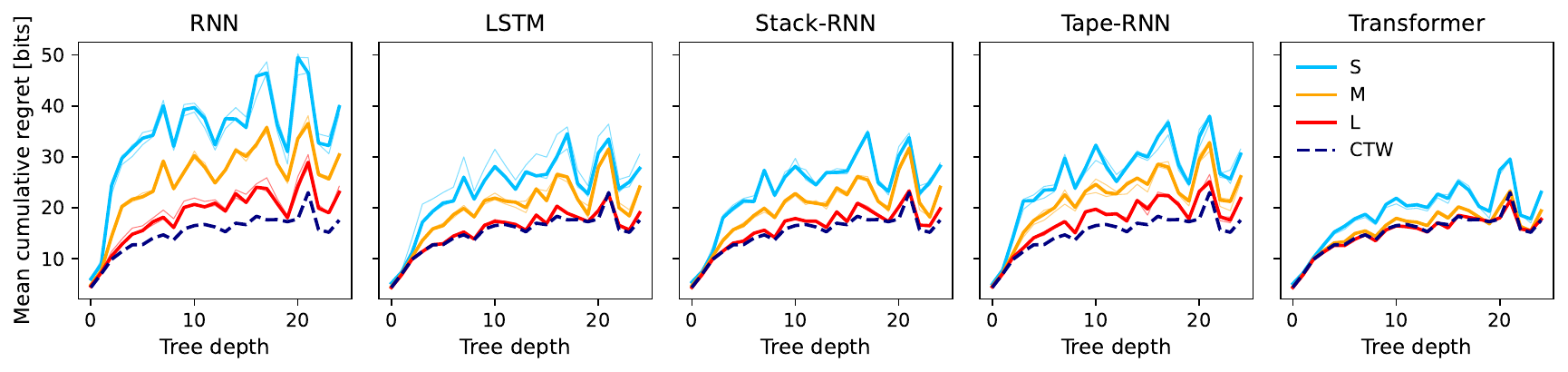}
        \caption{Average cumulative regret per tree depth of the generator.}
        \label{fig:redundancies_vs_tree_depth}
    \end{subfigure}
    \hfill 
    \begin{subfigure}[b]{\textwidth}
            \centering
            \includegraphics[width=\textwidth]{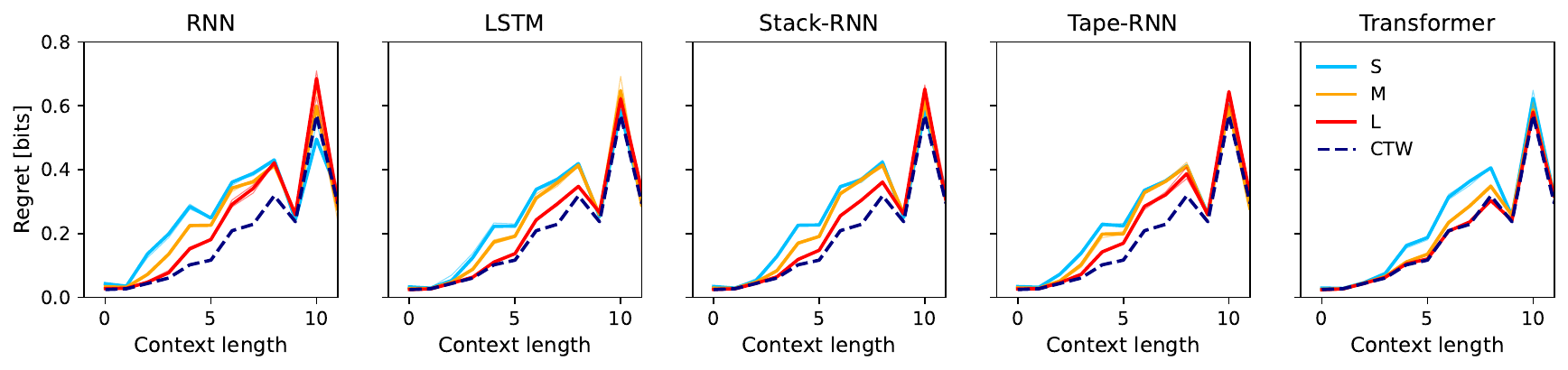}
            \caption{Average Instantaneous regret per current context length (only context-lengths up to $11$).}
        \label{fig:redundancies_vs_context_length}
        \end{subfigure}
    \begin{subfigure}[b]{\textwidth}
        \centering
        \includegraphics[width=\textwidth]{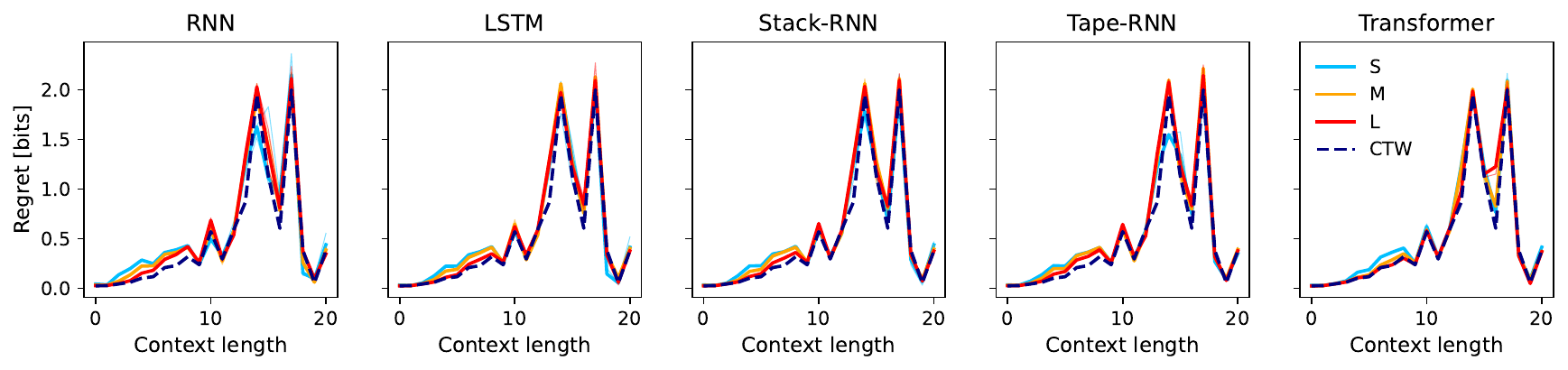}
        \caption{Average Instantaneous regret per current context length (all context-lenghts).}
        \label{fig:redundancies_vs_context_length_full}
    \end{subfigure}
    \caption{Detailed results for the same $6$k sequences as in Figure~\ref{fig:ctw_results}. %\ref{fig:ctw_redundancies}.
    Top two panels show histograms over tree depth (for all trajectories) and current context length (over all datapoints of all trajectories) use for evaluation in Figure~\ref{fig:ctw_results}. As expected, most generated trees have low depth and most datapoints have short contexts. The three lower panels show average cumulative regret per tree depth, and average instantaneous regret per context length respectively. Thin lines correspond to individual models (with different random initialization), bold lines show the median per model size. Across architectures smaller models only predict well for very short tree depth or very short context lengths (the maximum context length is upper bounded by the tree depth, but many contexts are much shorter than the maximum tree depth). Context lenghts~$\geq 11$ are rare, which makes quantitative results in this regime less reliable.}
    \label{fig:redundancies_vs_CTW}
\end{figure*}

\begin{figure*}[ht!]
    \centering
    \begin{subfigure}[b]{\textwidth}
        \centering
        \includegraphics[width=\textwidth]{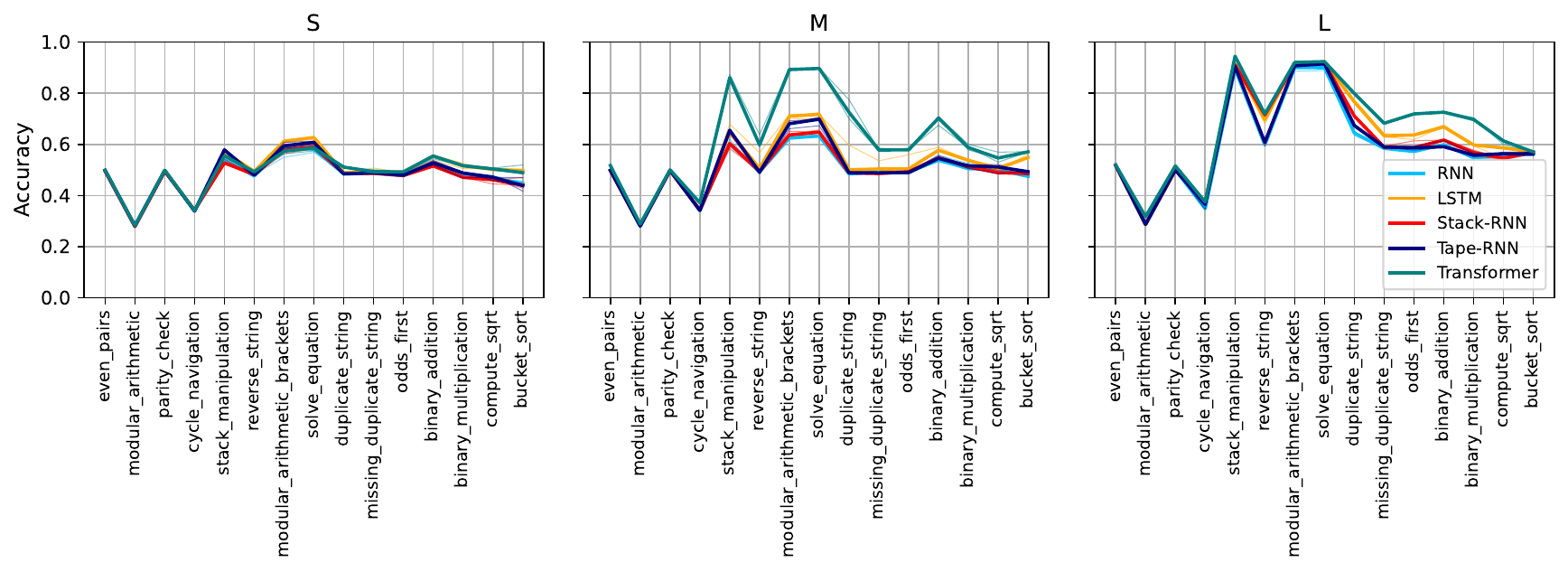}
        \caption{Mean accuracy per Chomsky task, grouped by model size.}
        \label{fig:chomsky_acc_vs_task}
        \hfill    
        \begin{subfigure}[b]{\textwidth}
            \centering
            \includegraphics[width=\textwidth]{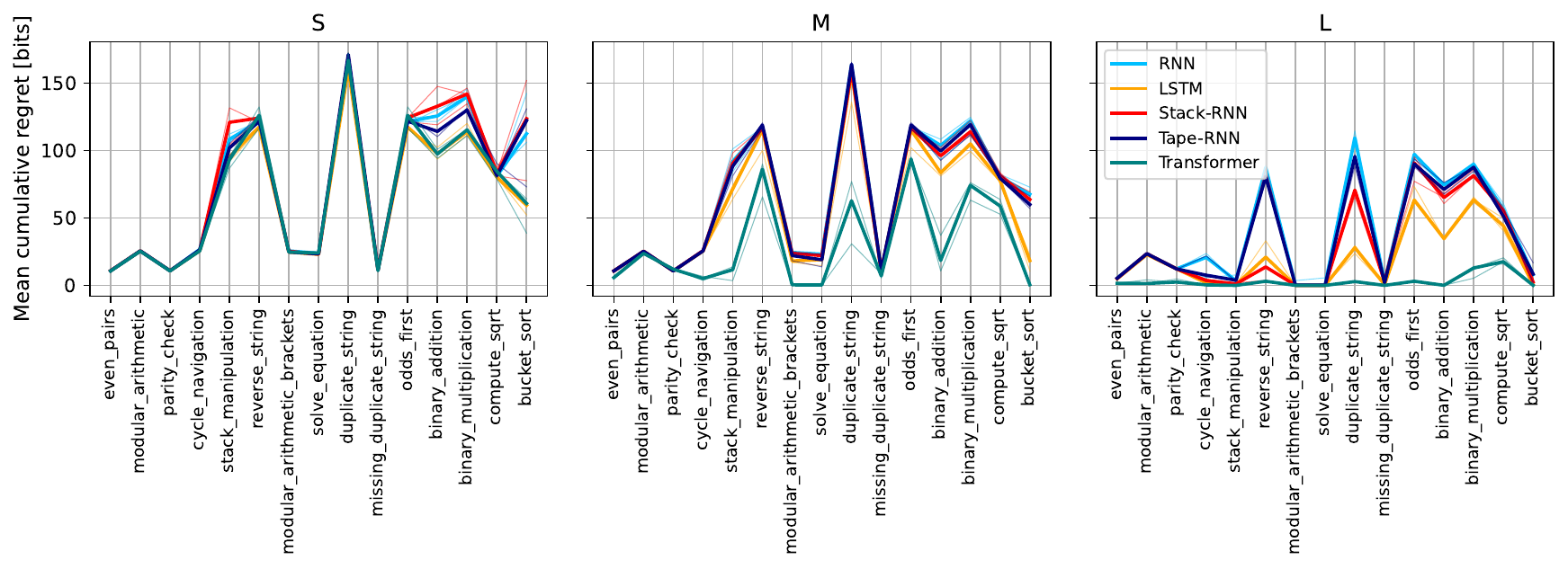}
            \caption{Mean cumulative regret per Chomsky task, grouped by model size.}
        \label{fig:chomsky_regret_vs_task}
        \end{subfigure}
    \end{subfigure}
    \caption{Detailed performance of networks trained and evaluated on the Chomsky tasks ($6$k sequences, $400$ sequences per task; main results shown in Figure~\ref{fig:main_results_chomsky}). Thin lines correspond to a single random initialization of a model, bolt lines show the median respectively.}
    \label{fig:chomsky_results_by_task}
\end{figure*}

\begin{figure*}[ht!]
    \centering
    \begin{subfigure}[b]{\textwidth}
        \centering
        \includegraphics[width=\textwidth]{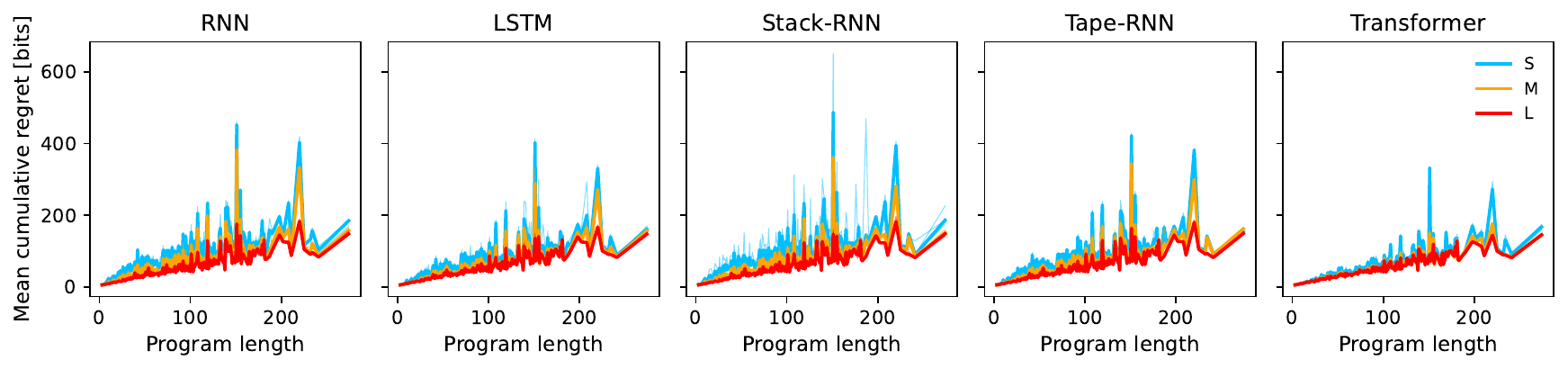}
        \caption{Average cumulative regret per program length.}
        % \label{fig:redundancies_vs_tree_depth}
        \hfill    
        \begin{subfigure}[b]{\textwidth}
            \centering
            \includegraphics[width=\textwidth]{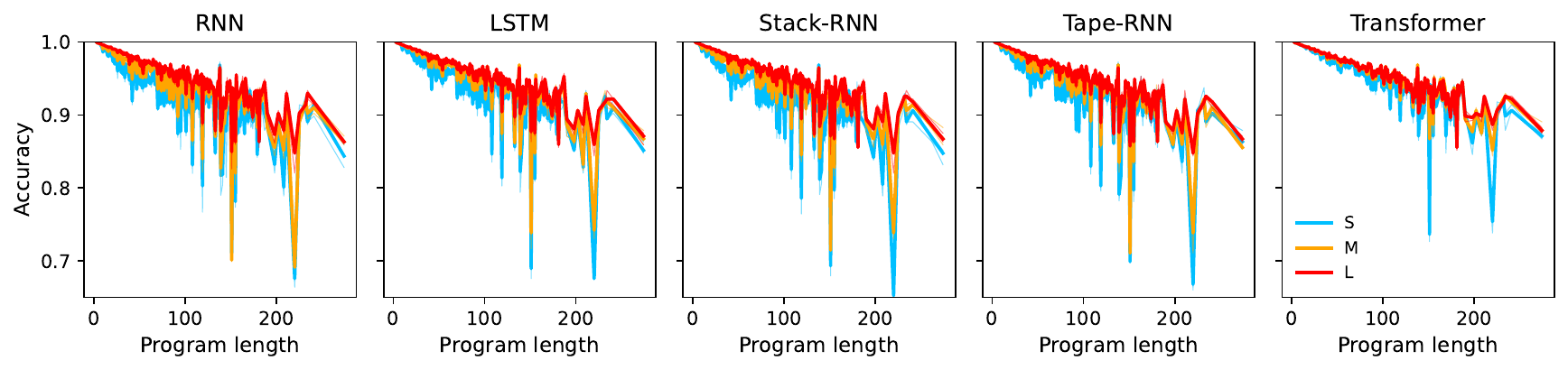}
            \caption{Average accuracy per program length.}
        % \label{fig:redundancies_vs_context_length}
        \end{subfigure}
    \end{subfigure}
    \hfill
    \begin{subfigure}[b]{0.27\textwidth}
        \centering
        \includegraphics[width=\textwidth]{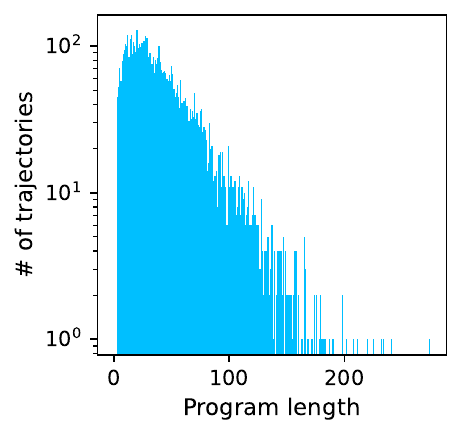}
        \caption{Histogram over program lengths.}
    \end{subfigure}
    \caption{Results per program length for UTM in-distribution evaluation (same data as in Figure~\ref{fig:main_results_utm}; $6$k sequences, length~$256$).}
    \label{fig:results_per_program_length_utm}
\end{figure*}

\begin{figure*}
    \centering
    \includegraphics[width=\textwidth]{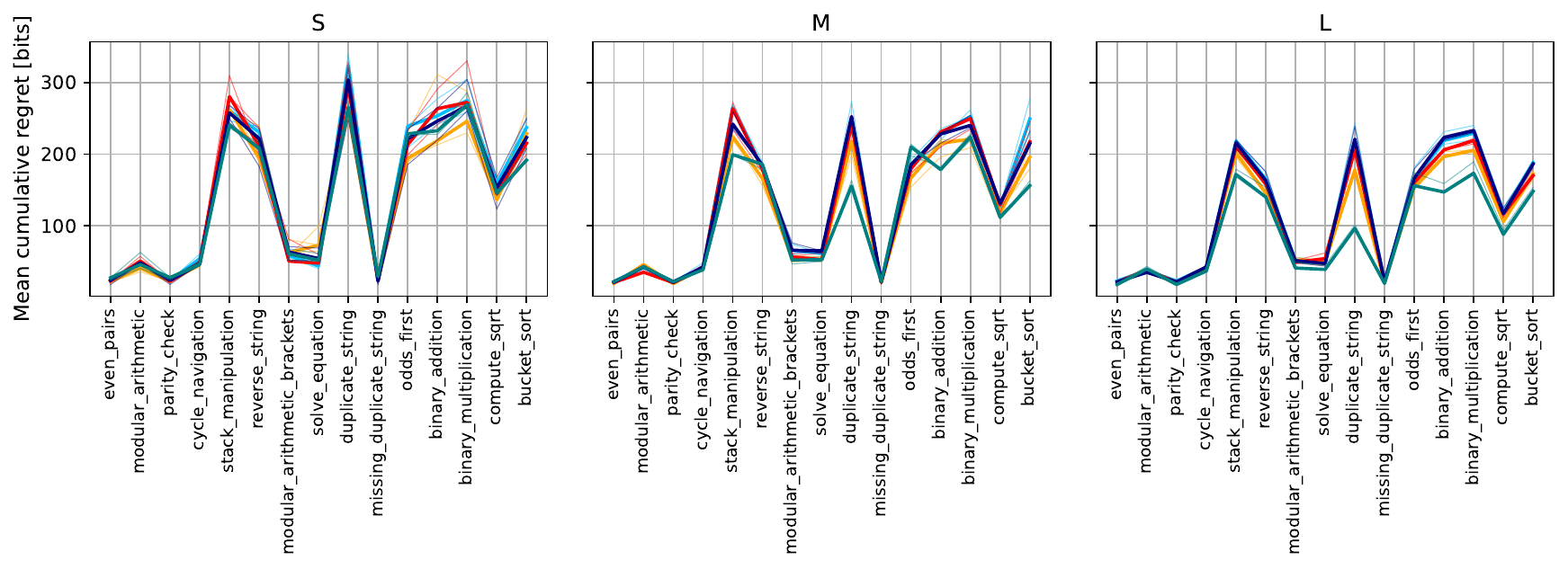}
    \includegraphics[width=\textwidth]{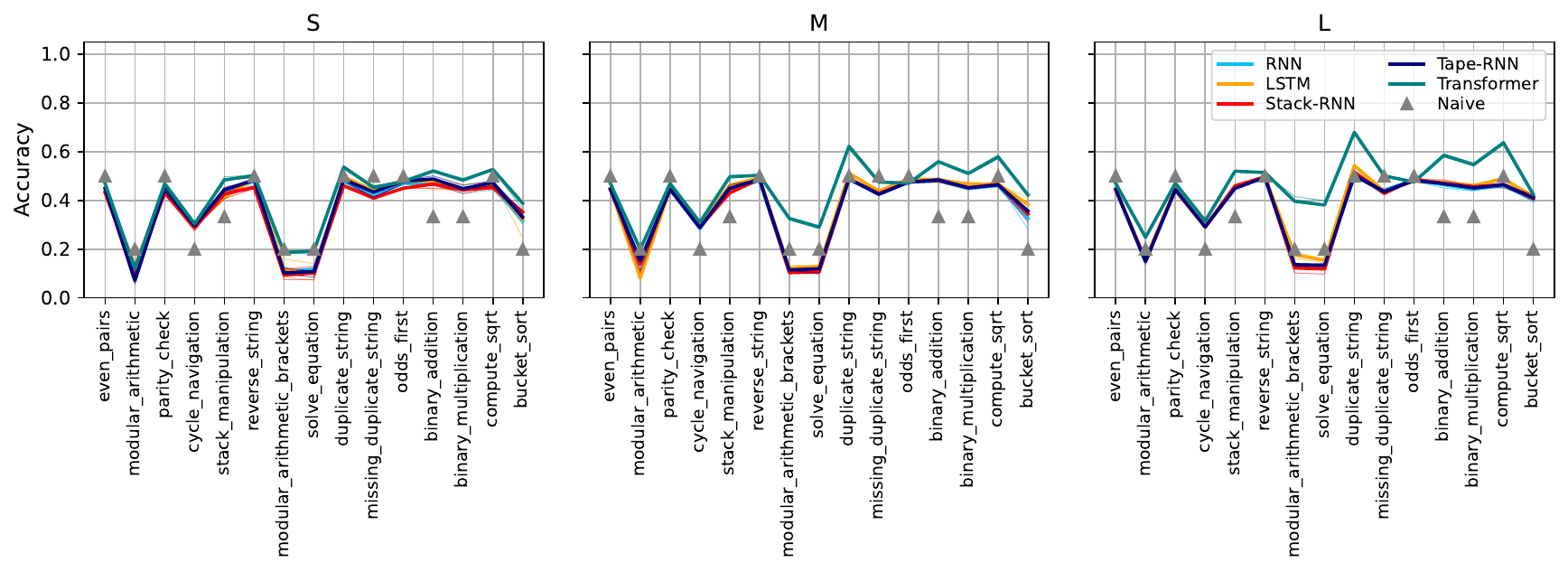}
    \caption{UTM transfer to Chomsky tasks.}
    \label{fig:utm_transfer_per_task}
\end{figure*}

\begin{figure*}[t]
    % \centering
    \begin{subfigure}[b]{\textwidth}
        \centering
        \includegraphics[width=0.9\textwidth]{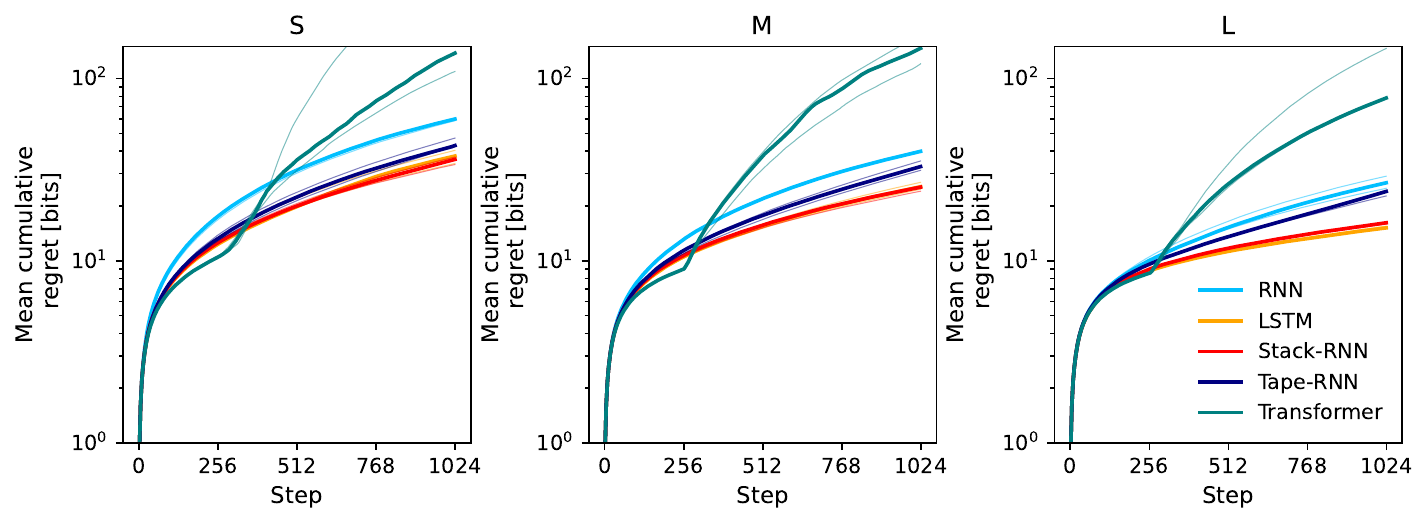}
        \caption{Variable-order Markov source (CTW) data.}
    \end{subfigure}
    % \hspace{1pt}
    \hfill
    \begin{subfigure}[b]{\textwidth}
        \centering
        \includegraphics[width=0.9\textwidth]{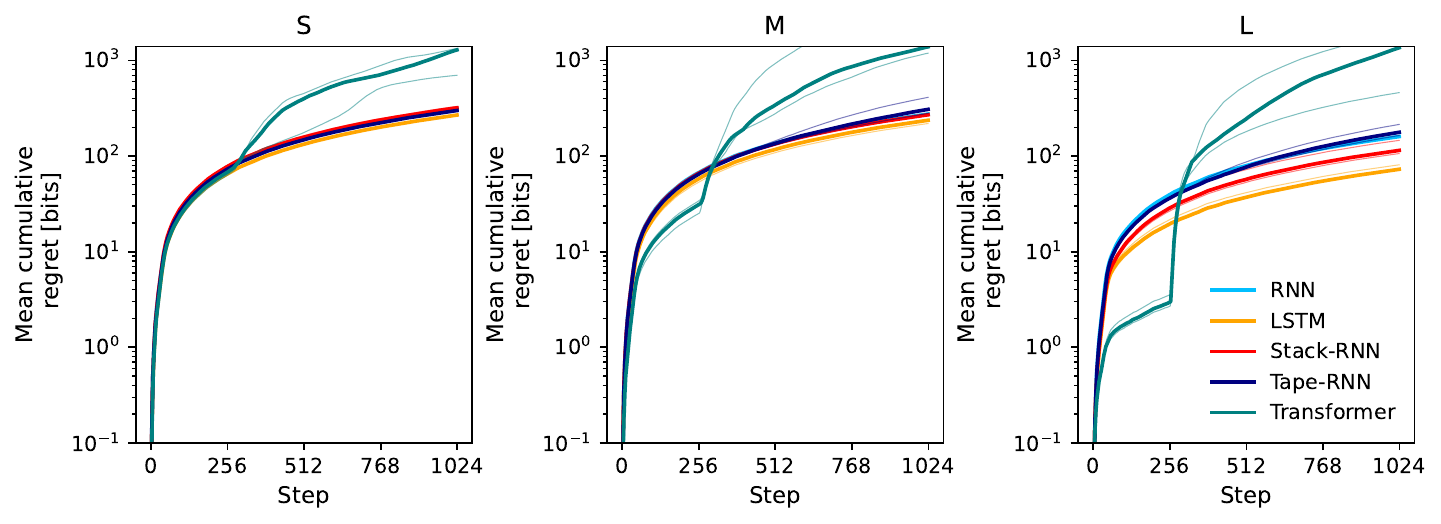}
        \caption{Chomsky tasks.}
    \end{subfigure}
    \hfill
    \begin{subfigure}[b]{\textwidth}
        \centering
        \includegraphics[width=0.9\textwidth]{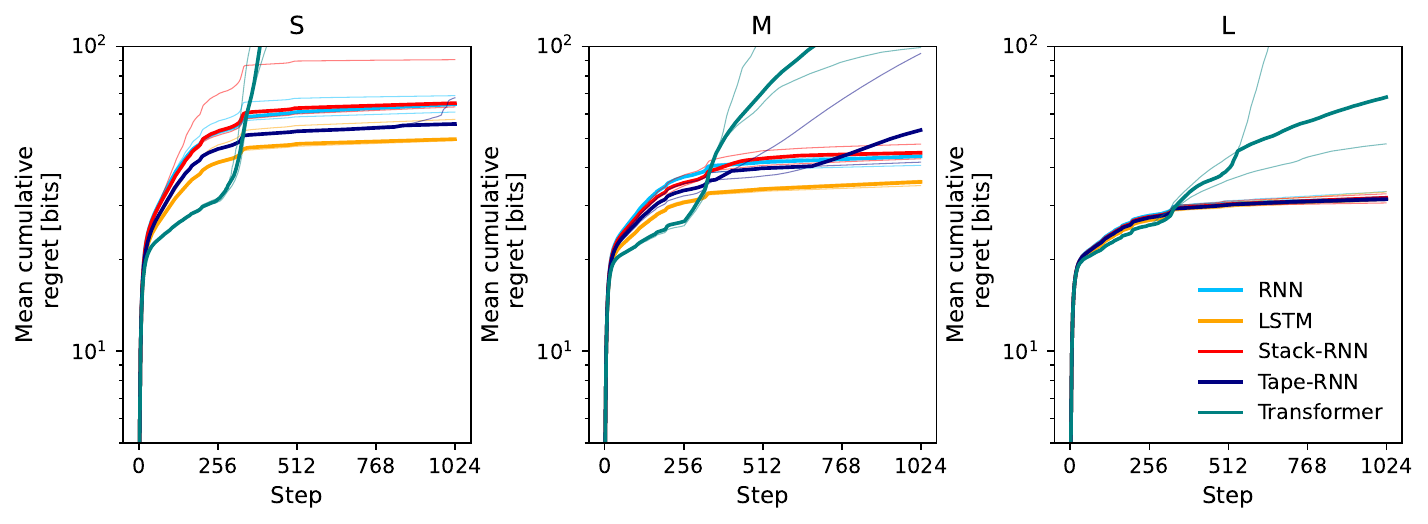}
        \caption{UTM data.}
    \end{subfigure}
    \caption{Full details of sequence-length generalization results. Models were trained on sequences of length~$256$ on their respective tasks, and are evaluated on $6$k sequences of length~$1024$ from the same data generator type. Thin lines show individual models, bold lines are the median across random initializations of the same model. As expected, all models perform fairly well up to their trained sequence length, and then performance deteriorates more or less sharply. Most notably, prediction performance of the transformer models, regardless of their size, degrades very rapidly after step~$256$ and is often an order of magnitude worse than the other models. Across all experiments, LSTMs perform best in terms of generalizing to longer sequences.}
    \label{fig:length_generalization_details}
\end{figure*}

Below we show additional results of the experiments on  the VOMS (Figure~\ref{fig:redundancies_vs_CTW}), the Chomsky  tasks (Figure~\ref{fig:chomsky_results_by_task}) and UTM source (Figures~\ref{fig:results_per_program_length_utm} and~\ref{fig:utm_transfer_per_task}). Finally, on Figure~\ref{fig:length_generalization_details} we show further details of the length generalization analysis.
%-------------------------------------------------------------%
\label{sec:ctw_result_details}

%-------------------------------------------------------------%
%\subsection{Chomsky Hierarchy Additional Results}
%-------------------------------------------------------------%

%-------------------------------------------------------------%
%\subsection{UTM Additional Results}
%-------------------------------------------------------------%

%-------------------------------------------------------------%
%\subsection{Sequence-Length Generalization Additional Results}
%-------------------------------------------------------------%

\end{document}